\documentclass[sigconf]{acmart}

\AtBeginDocument{%
  }

\setcopyright{acmlicensed}
\copyrightyear{2025}
\acmYear{2025}
\acmDOI{XXXXXXX.XXXXXXX}

\acmConference[Conference acronym 'XX]{Make sure to enter the correct
  conference title from your rights confirmation email}{June 03--05,
  2025}{Woodstock, NY}

\acmISBN{978-1-4503-XXXX-X/2018/06}

\usepackage{graphicx}
\usepackage{enumitem}
\usepackage{tcolorbox}
\usepackage{prettyref}
\usepackage{multirow,array}
\usepackage{listings}
\usepackage{stmaryrd} %for Iverson bracket
\usepackage{wrapfig}
\usepackage{subcaption}
\usepackage{tikz}
\usepackage{amsmath}
\usepackage[linesnumbered,ruled,vlined]{algorithm2e}
\usepackage{algorithmic}
\usepackage{arydshln}

\usepackage{pifont} 
\usepackage{rotating}

\newcommand{\cmark}{{\color{green!70!black}\ding{51}}} 
\newcommand{\xmark}{{\color{red!80!black}\ding{55}}}  

\makeatletter 
\g@addto@macro{\@algocf@init}{\SetKwInOut{Parameter}{Parameters}} 
\let\oldnl\nl% Store \nl in \oldnl
\newcommand{\nonl}{\renewcommand{\nl}{\let\nl\oldnl}}% Remove line number for one line
\makeatother
\SetKw{Continue}{continue\;}
\usepackage[normalem]{ulem}

\newtheorem{problem}{Problem}
\newtheorem{definition}{Definition}
\newtheorem{example}{Example}
\newtheorem{theorem}{Theorem}
\newtheorem{lemma}{Lemma}

\newcommand\themis{\textsc{Themis}}
\newcommand\aft{\textsc{Aft}}
\newcommand\sg{\textsc{Sg}}
\newcommand\vbt{\textsc{Vbt}}

\newcommand\expga{ExpGA}
\newcommand\cegarmethod{\textsc{CegarQuant}}
\newcommand\boxmethod{\textsc{BoxQTE}}
\newcommand\supp{Supplementary Material}

\newcommand\epsr{r_\epsilon}

\makeatletter
\protected\def\myDot{%
 \@ifnextchar.{}{.%
  \@ifnextchar,{}{%
   \@ifnextchar:{}{%
    \@ifnextchar;{}{%
     \@ifnextchar~{}{\ }
}}}}}

\newcommand\etal{{et~al}\myDot}
\newcommand\ie{i.\,e\myDot}

\newcommand\aka{a.\,k.\,a\myDot}
\renewcommand\And{\mathrel\wedge}
\newcommand\Or{\mathrel\vee}

\definecolor{newred}{RGB}{230,2,0}

\definecolor{newgray}{RGB}{230,230,230}

\newcommand\allsol{\text{Sol}}
\newcommand\Tool[1]{\textsc{#1}}

\newrefformat{sec}{Section~\ref{#1}}
\newrefformat{cha}{Chapter~\ref{#1}}
\newrefformat{app}{Appendix~\ref{#1}}
\newrefformat{def}{Definition~\ref{#1}}
\newrefformat{item}{\ref{#1}}
\newrefformat{tab}{Table~\ref{#1}}
\newrefformat{cor}{Corollary~\ref{#1}}
\newrefformat{ex}{Example~\ref{#1}}
\newrefformat{prop}{Proposition~\ref{#1}}
\newrefformat{fig}{Figure~\ref{#1}}
\newrefformat{alg}{Algorithm~\ref{#1}}
\newrefformat{line}{line~\ref{#1}}
\begin{document}

\title{Quantitative Verification of Fairness in Tree Ensembles}

\author{Zhenjiang Zhao}
\affiliation{%
  \institution{University of Electro-Communications}
  \city{Tokyo}
  %\state{}
  \country{Japan}
}
\email{zhenjiang@disc.lab.uec.ac.jp}

\author{Takahisa Toda}
\affiliation{%
  \institution{University of Electro-Communications}
  \city{Tokyo}
  %\state{}
  \country{Japan}
}
\email{toda@disc.lab.uec.ac.jp}

\author{Takashi Kitamura}
\affiliation{%
  \institution{Nat. Inst. of Advanced Industrial Science and Technology (AIST)}
  %\institution{National Institute of Advanced Industrial Science and Technology}
  \city{Tokyo}
  %\state{}
  \country{Japan}
}
\email{t.kitamura@aist.go.jp}

\renewcommand{\shortauthors}{Zhao et al.}

\begin{abstract}
  This work focuses on quantitative verification of fairness in tree ensembles.
  Unlike traditional verification approaches that merely return a single counterexample when the fairness is violated, quantitative verification estimates the ratio of all counterexamples and characterizes the regions where they occur, which is important information for diagnosing and mitigating bias.
  To date, quantitative verification has been explored almost exclusively for deep neural networks (DNNs). Representative methods, such as DeepGemini and FairQuant, all build on the core idea of Counterexample-Guided Abstraction Refinement, a generic framework that could be adapted to other model classes. We extended the framework into a model-agnostic form, but discovered two limitations: (i) it can provide only lower bounds, and (ii) its performance scales poorly.
  Exploiting the discrete structure of tree ensembles, our work proposes an efficient quantification technique that delivers any-time upper and lower bounds.
  Experiments on five widely used datasets demonstrate its effectiveness and efficiency.
  When applied to fairness testing, our quantification method significantly outperforms state-of-the-art testing techniques.
\end{abstract}




\maketitle

\section{Introduction}
\label{sec:introduction}
Machine Learning (ML)-based systems have been widely deployed, yet their growing societal influence poses significant risks and concerns, especially in safety-critical applications where correctness and reliability are essential.
There is an increasing interest in developing methods to verify whether such systems satisfy specific properties~\cite{Sanjit+CACM22}. 
Verification techniques aim either to provide a formal proof that a system adheres to a given property or to identify one counterexample demonstrating the violation.

Fairness and robustness are two important properties in the verification of ML-based systems.
\citet{SzegedyZSBEGF13} introduced the concept of \emph{adversarial examples} in image classification, demonstrating that imperceptible changes to an image could alter its classification. This discovery stimulated widespread interest in ML robustness~\cite{IJgoodICLR15,pmlr-v124-george-john20a}.
\emph{Local robustness} requires that for a given input, small perturbations to it do not change its classification outcome~\cite{IJgoodICLR15,7546524,7467366,7958570,10.5555/3157382.3157391,madry2019deeplearningmodelsresistant,Guan+CAV21}.
As for fairness, in recent years, some ML-based systems have been found to exhibit biases.
An example is COMPAS, an ML system employed in US courts, which has been shown to incorrectly predict that African American offenders have a higher risk of recidivism compared to white defendants~\cite{Angwin+2016}.
Such biased behaviors in ML systems have drawn increasing attention to ML fairness~\cite{Kitamura+SSBSE22,Pessach+ACM23,fairness+survey+Caton+ACM24,Chen+TOSEM24,zhao+ASE24,zhao2023ist}.
\emph{Global robustness}, a natural generalization of local robustness, requires that perturbations do not affect classification outcomes for any input across the entire input space~\cite{pmlr-v139-leino21a,pmlr-v124-george-john20a}.
The core idea of fairness is to ensure similar predictions for similar individuals. From this perspective, global robustness can be seen as a variant of fairness, as also suggested by \citet{Athavale+CAV24} and \citet{pmlr-v124-george-john20a}. In this work, we adopt the same viewpoint and consider global robustness as a form of fairness.

This study focuses on \emph{confidence-based fairness} (or  \emph{$\epsilon$-$\kappa$-fairness}) and \emph{confidence-based global robustness}  (or \emph{$\epsilon$-$\kappa$-robustness}), recently introduced by~\citet{Athavale+CAV24}.
These two properties generalize traditional fairness and robustness by incorporating a tolerance threshold $\epsilon$ and a confidence threshold $\kappa$.
The $\epsilon$ parameter allows comparisons between individuals who differ slightly in non-sensitive attributes, addressing the limitation of classical fairness definitions that only consider strictly identical inputs. Meanwhile, the $\kappa$ threshold ensures that only predictions with sufficiently high confidence are evaluated for fairness or robustness violations. This design reflects practical considerations, as low-confidence outputs are often excluded from decision-making processes in real-world applications.
The introduction of $\kappa$ also relaxes a strict requirement of global robustness and fairness with $\epsilon$, which implicitly assumes that decision boundaries cannot depend on continuous attributes, otherwise, counterexamples are inevitable.
\citet{Athavale+CAV24} also proposed a Marabou-based~\cite{marabou+CAV19} verification method for Deep Neural Networks (DNNs) to verify these confidence-based properties.

\emph{Quantitative verification} (or \emph{quantification} for short) plays an important role in understanding the overall behavior of ML systems.
While traditional verification\footnote{In the remainder of this paper, for the sake of clarity, we also use verification to denote traditional verification, and quantification to refer to quantitative verification.} techniques can formally prove the satisfaction of properties, they provide only a single counterexample when the violation occurs. 
Thus, verification techniques cannot answer quantification questions, such as the ratio of counterexamples over the entire input space.
Moreover, they are unable to capture the full counterexample space, limiting the insight into how and where properties fail.
Such quantitative information can be valuable to developers when evaluating or improving ML models, especially given the widespread presence of ML-based systems that violate these properties (e.g., most of the ML systems used as experimental benchmarks in many studies are unfair~\cite{Fan+ICSE22,fairify+biswas+ICSE23,fairquant+ICSE25}).

\emph{Tree ensembles}, such as Gradient Boosted Decision Trees (GBDTs)~\cite{chen+KDD16,Ke+NISP17,Bengio+NISP18} and Random Forests (RFs)~\cite{Breiman+ML01} are important ML models whose fairness-related properties deserve thorough investigation.
Although most existing work on fairness verification and testing has focused on DNNs, we argue that studying fairness in tree ensembles is also essential.
On the one hand, tree ensembles are popular and considered the preferred choice for solving tabular data problems among practitioners and in data science competitions,
as studies have shown that tree ensembles outperform DNNs on tabular data~\cite{Kossen+NISP21,Grinsztajn+NeurIPS22}.
On the other hand, fairness-related properties are crucial for tree ensembles, as they are widely used to handle problems involving individuals. This is because individuals are typically represented as tabular data, which tree ensembles are good at.

It is critical to develop quantification techniques for tree ensembles to handle $\epsilon$-$\kappa$-fairness and $\epsilon$-$\kappa$-robustness, given the importance of quantitative verification, the central role of tree ensembles in fairness-sensitive applications, and the practical relevance of confidence-based fairness and global robustness.
Although no quantification method currently exists for tree ensembles, several approaches have been proposed for DNNs, including DeepGemini~\cite{DeepGemini+AAAI23} and FairQuant~\cite{fairquant+ICSE25} (altough they cannot handle $\kappa$).
Both methods are based on the core idea of \emph{Counterexample-Guided Abstraction Refinement} (or \emph{CEGAR})\footnote{In the work of DeepGemini~\cite{DeepGemini+AAAI23}, it is called Counterexample-Guided Fairness Analysis (CEGFA).}, a general framework that can, in principle, be adapted to other model classes beyond DNNs.
In this work, we implement a CEGAR-based quantification method by generalizing the core idea of DeepGemini and FairQuant.

However, this CEGAR-based method has limitations:
(1) it provides only a lower bound on the quantification measure, and
(2) it suffers from scalability challenges due to the iterative refinement process.
To overcome these limitations, we propose a novel quantification technique tailored to tree ensembles, which leverages the discrete structure of their decision boundaries.
Unlike DNNs, whose decision boundaries are continuous, tree ensembles partition the input space into a finite set of \emph{hyperrectangles} (or \emph{boxes}).
By exploiting this structural nature, our approach enables more effective and efficient quantification while providing both upper and lower bounds.

\begin{table}[t]
\caption{Comparison of our work with existing methods.}
\label{tab:our_work_vs_others}
\centering
\renewcommand{\arraystretch}{1.25}
\resizebox{1\columnwidth}{!}{
\begin{tabular}{c|c|c|c|c}
\hline 
ML Model  & \multicolumn{2}{c|}{Verification} & \multicolumn{2}{c}{Quantification}\tabularnewline
\cline{2-5} \cline{3-5} \cline{4-5} \cline{5-5} 
 & Without Confidence  & Confidence-based  & Without Confidence  & Confidence-based\tabularnewline
\hline 
\multirow{4}{*}{DNNs} & Athavale~\etal~\cite{Athavale+CAV24},  & \multirow{4}{*}{Athavale~\etal~\cite{Athavale+CAV24}} &  Yang~\etal~\cite{Yang+TACAS21}, & \multirow{4}{*}{None}\tabularnewline
 & Fairify~\cite{fairify+biswas+ICSE23},  &  & DeepGemini~\cite{DeepGemini+AAAI23},  & \tabularnewline
 & DeepGemini~\cite{DeepGemini+AAAI23},  &  & Zhang~\etal~\cite{Zhang+TOSEM23},   & \tabularnewline
 & FairQuant~\cite{fairquant+ICSE25}  &  & FairQuant~\cite{fairquant+ICSE25} & \tabularnewline
\hline 
 & \textbf{Handled by this work},  & \multirow{4}{*}{\textbf{Handled by this work}} & \multirow{4}{*}{\textbf{Only this work}} & \multirow{4}{*}{\textbf{Only this work}}\tabularnewline
Tree & Kantchelian~\etal~\cite{Kantchelian+ICML16},  &  &  & \tabularnewline
Ensembles  & Chen~\etal~\cite{chen+NISP19},  &  &  & \tabularnewline
 & VERITAS~\cite{veritas+Laurens+ICML21}  &  &  & \tabularnewline
\hline 
\end{tabular}
}
\end{table}

\paragraph{Our contributions.}
In summary, the key contributions of this work are listed as follows.
\begin{itemize}[label=$\bullet$]
    \item We provide a formal definition for quantitative verification confidence-based global robustness and fairness.
    \item We propose \boxmethod{}, a novel quantification method tailored to tree ensembles.
    \boxmethod{} supports parallel computation and can provide any-time lower and upper bounds for quantitative measures.
    Furthermore, we propose three enhancement techniques to enhance its performance.
    \item We implement \cegarmethod{} by generalizing the core idea behind DeepGemini and FairQuant, which can serve as a state-of-the-art quantification method baseline.
    \item We implement our method \boxmethod{}, and 
    conduct experimental evaluation, demonstrating that:
    (1) \boxmethod{} outperforms the state-of-the-art quantification method in both effectiveness and efficiency, (2)  the three enhancement techniques contribute to performance gains, and (3) when applied to fairness testing, \boxmethod{} surpasses the state-of-the-art fairness testing methods in efficiency.
\end{itemize}
\prettyref{tab:our_work_vs_others} summarizes the existing verification and quantification methods for robustness and fairness properties, illustrating the research gaps addressed by this work.
Notably, our quantification methods can also be applied when confidence is not a concern by adjusting the confidence threshold to a value that satisfies all inputs.
Moreover, although this work does not address traditional verification, quantification inherently subsumes it.

\section{Preliminary}
\label{sec:preliminary}

This study focuses on \emph{binary classification models} and tabular datasets. 
A binary classifier is defined as a function $f : X \to \{-1, 1\}$, mapping the input space $X$ to two classifications $-1$ (negative) and $1$ (positive). 
Each instance of a tabular dataset, to be input to the classifier $f$, consists of $n$ attributes, with their indices denoted by $A= \{1, \dots, n\}$.
The input space is represented as $X = X_1 \times \dots \times X_{n}$, where $X_i$ is the domain of the $i$-th attribute.
An input is denoted by $\vec{x} = (x_1, \dots, x_{n}) \in X$.
For each input $\vec{x}$, we denote the outcome of
prediction and prediction confidence of $f$ by $f(\vec{x})$ and $f_c(\vec{x})$, respectively.

\subsection{Fairness and Robustness}

We review the concepts of fairness and global robustness  with classifier confidence, the main target properties in this work, as introduced by \citet{Athavale+CAV24}.
Other basic definitions of robustness and fairness are summarized in \supp{}.

\begin{definition}[Global $\epsilon$-Robustness with Classifier Confidence $\kappa$: \aka, $\epsilon$-$\kappa$-Robustness]
Given a tuple of real numbers $\vec{\epsilon} = (\epsilon_1, \dots, \epsilon_n)$, where $\epsilon_i>0$, and a real number $\kappa$ such that $0 \leq \kappa < 1$, a classifier $f$ is \emph{globally $\epsilon$-$\kappa$-robust} if the following condition is satisfied:
\begin{equation*}
\begin{aligned}
\forall \vec{x}, \vec{x'} \in X, \bigwedge_{i=1}^{n} |x_i - x_i'| \leq \epsilon_i \land f_c(\vec{x}) > \kappa \to f(\vec{x}) = f(\vec{x'}).
\end{aligned}
\end{equation*}
\end{definition}

\begin{definition}[$\epsilon$-Fairness with Classifier Confidence $\kappa$: \aka, $\epsilon$-$\kappa$-Fairness]
Given a tuple of real numbers $\vec{\epsilon} = (\epsilon_1, \dots, \epsilon_n)$, where $\epsilon_i>0$, a real number $\kappa$ such that $0 \leq \kappa < 1$, and an index set of sensitive attributes $S \subseteq A$, a classifier $f$ is \emph{$\epsilon$-$\kappa$-fair} with respect to $S$ if the following condition is satisfied:
\begin{equation*}
\begin{aligned}
\forall \vec{x}, \vec{x'} \in X, & \bigwedge_{i \in A \setminus S} |x_i - x'_i| \leq \epsilon_i \land \bigvee_{j \in S} x_j \neq x'_j \land f_c(\vec{x}) > \kappa
\\ & \to f(\vec{x}) = f(\vec{x'}).
\end{aligned}
\end{equation*}
\end{definition}

\subsection{Decision Trees and Tree Ensembles}

\paragraph{Basics}
A tree ensemble classifier consists of multiple decision trees. 
A decision tree is composed of three types of nodes: The \emph{root node}, \emph{internal nodes}, and \emph{leaf nodes}.
The root or each internal node contains a \emph{split condition} and links to the left and a right child node. 
Each leaf node contains a real value, named \emph{leaf value}, and does not have any child nodes.
Given an input $\vec{x} \in X$, a decision tree $T_j$ is traversed recursively from the root node. At each node, if $\vec{x}$ satisfies the splitting condition, it moves to the left child node. Otherwise, it moves to the right.
The traversal ends when a leaf node is reached and the leaf value is returned  as the output of tree $T_j$ for input $\vec{x}$, denoted as $T_j(\vec{x})$.
A \emph{path} $\pi$ is a sequence of nodes from the root, passing through intermediate nodes, to a leaf. The leaf value corresponding to path $\pi$ is denoted as $\text{value}(\pi)$.

\paragraph{Box of path}
Following the formulation of Chen \etal~\cite{chen+NISP19}, we refer to the set of inputs characterized by a path $\pi$ as the \emph{box} of path $\pi$.
That is, the \emph{box} of $\pi$ is the set of all inputs that satisfy the conditions along the path and therefore traverse it. 
This set forms a \emph{hyperrectangle} (i.e., a Cartesian product of intervals) in the input space $X$, and we denote it as:
\begin{equation}
    B^{(\pi)} = [l^{(\pi)}_1, r^{(\pi)}_1) \times \dots \times [l^{(\pi)}_n, r^{(\pi)}_n),
\end{equation}
where $[l^{(\pi)}_i, r^{(\pi)}_i)$ is an interval showing the range of the $i$-th attribute, with $l^{(\pi)}_i$ and $r^{(\pi)}_i$ as its minimum and maximum bounds.
The box $B^{(\pi)}$ is derived from the split conditions of nodes in the path $\pi$ and the domain constraints of attributes.

\paragraph{Semantics of tree ensembles in terms of box}
We formalize the prediction process of a decision tree of a tree ensemble, using the 
concept of the boxes of paths.
Given an input $\vec{x} \in X$, the decision tree $T_j$ returns a leaf value, defined as:
\begin{equation}
    T_j(\vec{x}):=\sum_{\text{path}\, \pi\, \text{in}\, T_j} \llbracket  \vec{x} \in B^{(\pi)} \rrbracket \cdot \text{value}(\pi),
\end{equation}
where $\llbracket \cdot \rrbracket$ is the Iverson bracket, evaluating to $1$ if the condition holds, otherwise $0$.

\begin{figure}[t]
  \begin{center}
  \resizebox{1\columnwidth}{!}{
  \includegraphics{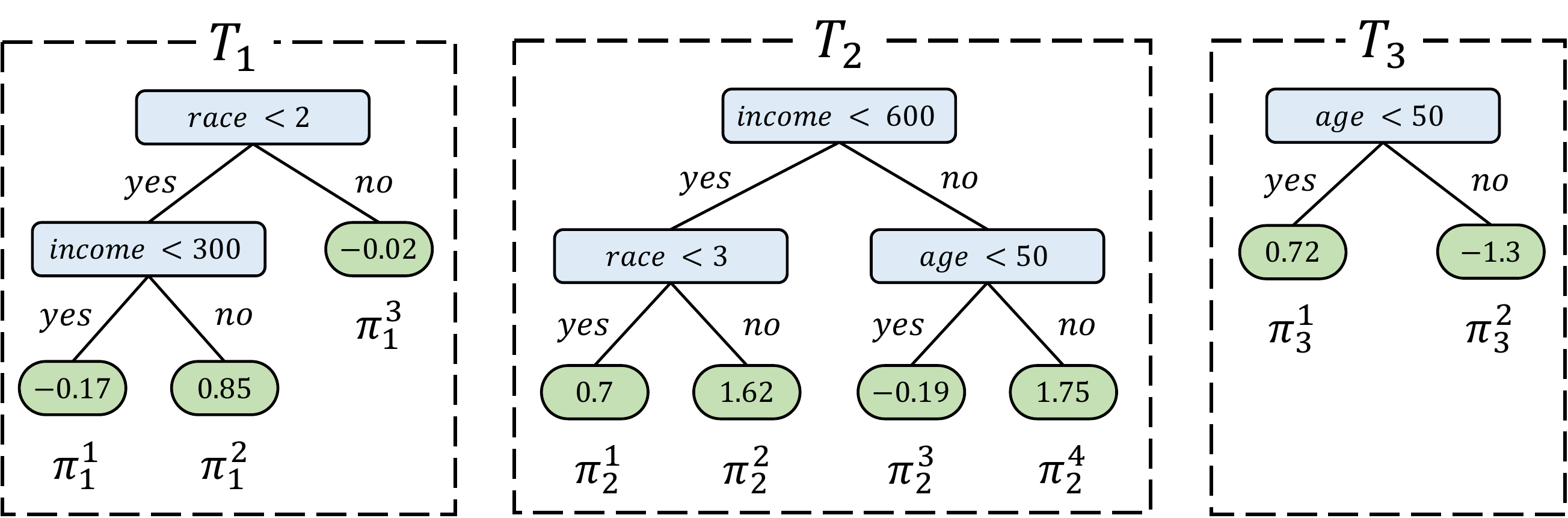}
  }
  \end{center}
    \caption{An example of GBDT tree ensemble classifier for predicting loan eligibility.
    The input attributes are ordered as `income', `race' and `age'.
    Among them, income takes real values over the range $[0, 1000]$, race takes categorical values over the range $[0, 4]$, and age takes integer values over the range $[0, 100]$.
    }
    \label{fig:tree-ensemble}
\end{figure}

\begin{example}[Tree Ensemble]
    \prettyref{fig:tree-ensemble} shows a tree ensemble consisting of three decision trees ($T_1$, $T_2$ and $T_3$). 
    For example, the first decision tree, $T_1$, has three paths $\pi_1^1$, $\pi_1^2$, and $\pi_1^3$; $T_2$ and $_3$ respectively have four and two paths.
    The boxes of the three paths of $T_1$ are respectively $B^{(\pi_1^1)}=[0,300)\times[0,2)\times[0,100]$, $B^{(\pi_1^2)}=[300,1000]\times[0,2)\times[0,100]$, and $B^{(\pi_1^3)}=[0,1000]\times[2,4]\times[0,100]$.
For the input $\vec{x} = (500, 1, 40)$, the tree $T_1$ follows the path $\pi_1^2$ and produces the output $T_1(\vec{x}) = 0.85$.
Tree $T_2$ follows the leftmost path and outputs $0.70$, while $T_3$ follows the left path and outputs $0.72$. 
\end{example}

Suppose a tree ensemble classifier $f$ consists of $m$ decision trees $T_1, \dots, T_m$. 
For an input $\vec{x} \in X$, the tree ensemble $f$ first computes the confidence of classifying $\vec{x}$ as classification $1$, denoted as $f_c^{(1)}$, by summing the outputs of the $m$ trees on $\vec{x}$.
If $f_c^{(1)} > \frac{1}{2}$, the input $\vec{x}$ is classified as classification $1$ (\ie, positive); otherwise, it is classified as $-1$ (\ie, negative).
The deisgn of $f_c^{(1)}$ depends on the tree ensemble type: for GBDTs, $f_c^{(1)}(\vec{x}):= \text{sigmoid}(\sum_j^m T_j(\vec{x}))$; for random forests, $f_c^{(1)}(\vec{x}):= \frac{1}{m}\sum_j^m T_j(\vec{x})$.
Formally, $f$ and its prediction confidence function $f_c$ can be defined as:
\begin{equation}
    f(\vec{x}) := \llbracket f_c^{(1)}  > \frac{1}{2} \rrbracket - \llbracket  f_c^{(1)}  < \frac{1}{2} \rrbracket,\ \text{and}\  
    f_c(\vec{x}) := \text{max}(f_c^{(1)}, 1-f_c^{(1)}).
\end{equation}

We also revisit some useful concepts of a \emph{path tuple} and its corresponding \emph{box}, as introduced by~\cite{chen+NISP19}, for analyzing tree ensembles.  
For a tree ensemble \( f \), a \emph{path tuple} \( \vec{\pi}(f) \) is a sequence of paths \( \vec{\pi}(f) = (\pi_1, \dots, \pi_m) \), where \( \pi_j \) is a path in the \( j \)-th tree \( T_j \).  
When the context is clear, we write \( \vec{\pi}(f) \) simply as \( \vec{\pi} \).  
The \emph{box} of a path tuple \( \vec{\pi} \), denoted \( B^{(\vec{\pi})} \), is the intersection of the boxes of its path:
$B^{(\vec{\pi})} = \bigcap_{j=1}^m B^{(\pi_j)}.$
Since the intersection of hyperrectangles remains a hyperrectangle, a box of a path tuple is also a hyperrectangle. We denote it as $B^{(\vec{\pi})} = [l^{(\vec{\pi})}_1, r^{(\vec{\pi})}_1) \times \dots \times [l^{(\vec{\pi})}_n, r^{(\vec{\pi})}_n)$, where $l^{(\vec{\pi})}_i$ and $r^{(\vec{\pi})}_i$ are minimum and maximum bounds of $i$-th attribute for the path tuple $\vec{\pi}$.

\begin{example}[Tree Ensemble]
    In \prettyref{fig:tree-ensemble}, the tree ensemble $f$ haves $3 \times 4 \times 2 = 24$ path tuples.
    For example, the box of path tuple $\vec{\pi} = (\pi^2_1, \pi^1_2, \pi^1_3)$ is $B^{(\vec{\pi})} = B^{(\pi^2_1)} \cap B^{(\pi^1_2)} \cap B^{(\pi^1_3)} = [300,600)\times[0,2)\times[0,50)$.
    For an input $\vec{x}=(500,1,40)$, $f$ follows the paths $\pi^2_1, \pi^1_2, \pi^1_3$. Since $\text{sigmoid}(T_1(\vec{x})+T_2(\vec{x})+T_3(\vec{x})) = \text{sigmoid}(2.27) \approx 0.91$ exceeds $\frac{1}{2}$, $\vec{x}$ is classified as classification $1$, with $f_c(\vec{x})\approx 0.91$.
\end{example}

\section{Methodology and Theory}
\label{sec:proposed-method}

This section details our proposed quantification method for tree ensembles, called \boxmethod{} (short for box-based quantification for tree ensembles).
For brevity, we focus on $\epsilon$-$\kappa$-fairness. The $\epsilon$-$\kappa$-robustness scenario can be handled by ignoring sensitive attributes.

\subsection{Quantitative Verification Problem}
\label{sec:quantitative-measure-hardness}

\citet{DeepGemini+AAAI23} defines a measure for quantifying individual fairness as follows.

\begin{definition}[Individual Fairness Measure]
The individual fairness measure of a classifier $f$, denoted by $\mathcal{M}_{\text{if}}$, is the proportion of inputs, among input space $X$, for which modifying sensitive attributes does not change the classification outcome of $f$.
\end{definition}

However, the $\epsilon$-$\kappa$-robustness and $\epsilon$-$\kappa$-fairness properties allow input variations within the tolerance range $\vec{\epsilon}$ and consider only inputs with a confidence level higher than the threshold $\kappa$.
Building upon the definition by \cite{DeepGemini+AAAI23}, we further introduce the following measures to evaluate $\epsilon$-$\kappa$-robustness and $\epsilon$-$\kappa$-fairness.

\begin{definition}[$\epsilon$-$\kappa$-Fairness Measure]
Given a classifier $f$, a tolerance $\vec{\epsilon}$, and a confidence threshold $\kappa$,
the $\epsilon$-$\kappa$-fairness measure of $f$, denoted by $\mathcal{M}_{\text{fair}}$, is the proportion of inputs, among those with prediction confidence level exceeding $\kappa$, for which modifying non-sensitive attributes within $\vec{\epsilon}$ and altering sensitive attributes do not change the classification outcome of $f$.
\end{definition}

\begin{definition}[$\epsilon$-$\kappa$-Robustness Measure]
Given a classifier $f$, a tolerance $\vec{\epsilon}$, and a confidence threshold $\kappa$,
the $\epsilon$-$\kappa$-robustness measure of $f$, denoted by $\mathcal{M}_{\text{robust}}$, is the proportion of inputs, among those with prediction confidence level exceeding $\kappa$, for which modifying attributes within $\vec{\epsilon}$ does not change the outcome of $f$.
\end{definition}

\begin{problem}[Quantitative Verification]
Calculate $\mathcal{M}_{\text{robust}}$ or $\mathcal{M}_{\text{fair}}$ for  a classifier $f$.
\end{problem}

\subsection{Theory Basics}
\label{sec:theoretical_gurantees}

Our proposed quantifier leverages nature of tree ensembles: 
(1) the union of path-tuple boxes covers the entire input space,
(2) path-tuple boxes are pairwise disjoint, and 
(3) all inputs within a box share the same prediction and confidence.

While this idea is implicit in the work of \citet{chen+NISP19}, it was not explicitly formalized. We formalize it in the following lemma.
\begin{lemma}
Let $f$ be a tree ensemble, and $X$ be its input space. 
The following statements hold: (1) $\bigcup_{\text{path tuple }\vec{\pi} \text{ in } f}B^{(\vec{\pi})} = X$,
(2) for any two path tuples $\vec{\pi}$ and $\vec{\pi'}$, if $\vec{\pi} \neq \vec{\pi'}$, then $B^{(\vec{\pi})} \cap B^{(\vec{\pi'})} = \emptyset$, and
(3) for any path tuple $\vec{\pi}$, $\forall \vec{x}, \vec{x'} \in B^{(\vec{\pi})}, f(\vec{x}) = f(\vec{x'}) \land f_c(\vec{x}) = f_c(\vec{x'})$.
\label{lemma:partitioned-by-boxes}
\end{lemma}

\begin{figure}[t]
  \centering
  \includegraphics[width=.95\columnwidth]
  {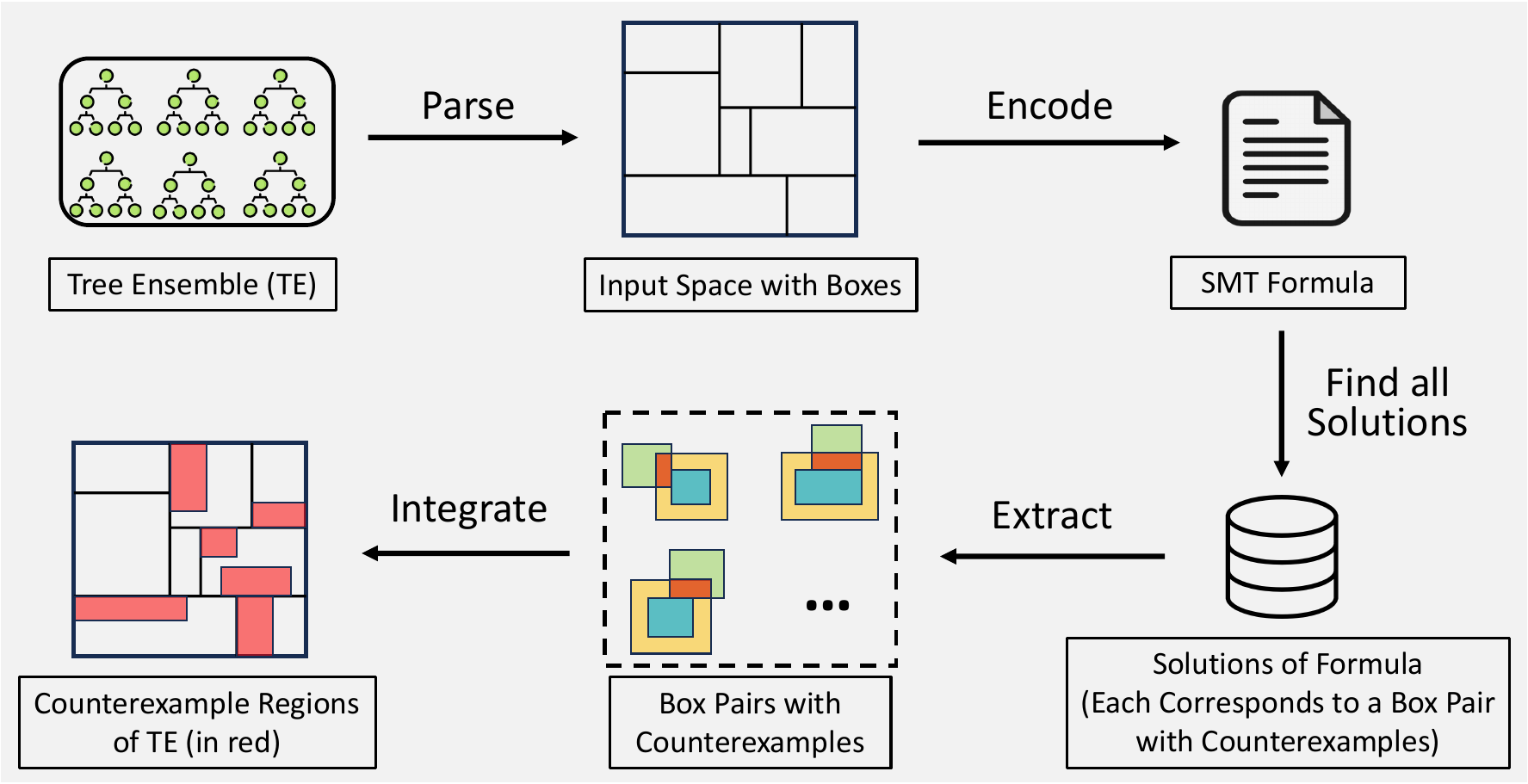}
  \caption{Overview of our quantification method, \boxmethod{}.}
  \label{fig:overview-of-our-method}
\end{figure}

\subsection{Intution of Our Quantification Method}

\prettyref{fig:overview-of-our-method} presents an overview of \boxmethod{}.
The core intuition is threefold:
(1) A tree ensemble partitions the input space into pairwise disjoint boxes.
(2) For any two boxes, we can construct the corresponding counterexample region.
(3) By inspecting every pair of boxes, we recover all counterexample regions.
To avoid brute-force search, we formulate an SMT formula whose solutions correspond exactly to the pairs of boxes that contain counterexamples.
\prettyref{sec:smt-encoding-for-TE} describes the SMT encoding, and \prettyref{sec:boxqte} elaborates on the algorithm together with its theoretical guarantees.

\subsection{SMT Encoding for Tree Ensembles}
\label{sec:smt-encoding-for-TE}

\lstset{stringstyle=\ttfamily,
        columns=fullflexible,
        backgroundcolor = \color{newgray},
        %frame = trBL,
        frame = single,
        %frame = lines,
        %basicstyle=\scriptsize,
        %basicstyle=\scriptsize\linespread{1.1}\selectfont,
        basicstyle=\scriptsize\linespread{1.32}\selectfont,
        breaklines=true,
        %postbreak=\mbox{\textcolor{red}{$\hookrightarrow$}\space},
        %numbers=left,
        %numberstyle=\tiny,
        numbersep= -7pt,
        morecomment=[l][\color{blue}]{//}
        %framexleftmargin=1em
}

We aim to construct an SMT formula whose satisfying assignments are counterexamples (i.e., input pairs that violate the target property), while also capturing the path tuples to which those inputs belong.
We list variables for encoding as follows:
\begin{itemize}[label=$\bullet$] 
    \item Input variables: $\vec{x} = (x_1, \dots, x_{n})$ and $\vec{x'} = (x'_1, \dots, x'_{n})$.%
    \item Tree value variables: $\vec{v} = (v_1, \dots, v_m)$ and $\vec{v'} = (v'_1, \dots, v'_m)$.
    \item Path variables: $\vec{p} = (p_1, \dots, p_m)$ and $\vec{p'} = (p'_1, \dots, p'_m)$.
\end{itemize}

The tree ensemble classifier with \( m \) trees over \( n \) attributes is encoded by sequentially encoding each node along every path in each tree, as follows\footnote{It is worth noting that a number of studies have addressed the encoding of tree ensembles, including approaches based on SMT and MILP. This work presents an intuitive SMT encoding that builds upon the encoding for decision trees by \citet{Sharma+ICTSS20} and 
\citet{zhao2023ist}.}:
\begin{equation}
\begin{aligned}
\Psi_{\text{trees}}(\vec{x}, \vec{v}, \vec{p}):= & \bigwedge_{j=1}^m \bigwedge_{\text{path } \pi  \textit{ in } T_j} \bigwedge_{i=1}^n l_i^{(\pi)} \leq x_i < r_i^{(\pi)} \\ & \to
(v_j = \text{value}(\pi)) \land (p_j = \text{id}(\pi)),\label{eq:tree-in-out}
\end{aligned}
\end{equation}
where $\text{value}(\pi)$ is the leaf value of the path $\pi$ and $\text{id}(\pi)$ gives the unique value for $\pi$.

We encode the classifier confidence level $\kappa$, using the property $\text{logit}(1-\kappa) = -\text{logit}(\kappa)$, as follows:
\begin{equation}
\begin{aligned}
\Psi_{\kappa}(\vec{v}, \kappa):= \sum_{j=1}^m v_j > \text{logit}(\kappa) \lor \sum_{j=1}^m v_j < -\text{logit}(\kappa),
\end{aligned}
\end{equation}
where $\text{logit}(\cdot)$ is the inverse of the sigmoid function.

The following constraints, together with Equation~\eqref{eq:tree-in-out}, ensure that \( \vec{x} \) and \( \vec{x'} \) receive different prediction outcomes:
\begin{equation}
\begin{aligned}
\Psi_{\text{ineq}}(\vec{v}, \vec{v'}) := \left( \sum_{j=1}^m v_j < 0 \land \sum_{j=1}^m v_j' > 0 \right) \lor \left(\sum_{j=1}^m v_j > 0 \land \sum_{j=1}^m v_j' < 0 \right).
\end{aligned}
\end{equation}

The following constraints ensure that an input pair $(\vec{x},\vec{x'})$ does not satisfy fairness under tolerance $\vec{\epsilon}$:
\begin{equation}
\begin{aligned}
\Psi_{\text{unfair}}(\vec{x}, \vec{x'}, \vec{v}, \vec{v'}, \vec{\epsilon}): = 
\bigwedge_{i \in  A \setminus S} |x_i - x_i'| < \epsilon_i 
\land \bigvee_{j \in S} x_j \neq x_j' 
\land \Psi_{\text{ineq}}(\vec{v}, \vec{v'}),
\end{aligned}
\end{equation}
where note that $|x_i - x_i'|< \epsilon_i$ can be encoded as $(-\epsilon_i < x_i - x_i') \land (x_i - x_i' < \epsilon_i)$.

Using the above encoding, we construct SMT formula 
$\Phi_{\text{unfair}}(\epsilon, \kappa)$, which are abbreviated as $\Phi_{\text{unfair}}$ when the context is clear, 
for verifying $\epsilon$-$\kappa$-fairness, as follows\footnote{ \supp{} shows an example of SMT formula for GBDT in \prettyref{fig:tree-ensemble}.}:  
\begin{equation}
\begin{aligned}
\Phi_{\text{unfair}}(\vec{\epsilon}, \kappa): = & \Psi_{\text{trees}}(\vec{x}, \vec{v}, \vec{p}) \land \Psi_{\text{trees}}(\vec{x'}, \vec{v'}, \vec{p'}) \land \Psi_{\kappa}(\vec{v}, \kappa) \\& \land  \Psi_{\text{unfair}}(\vec{x}, \vec{x'}, \vec{v}, \vec{v'}, \vec{\epsilon}).\label{F_fair}
\end{aligned}
\end{equation}

It is evident that the solutions of $\Phi_{\text{unfair}}$ correspond one-to-one to the input pairs of $f$ that violate fairness.
Please refer to the Appendix for an example of the SMT encoding for a tree ensemble.

\subsection{Proposed Quantification Method: \boxmethod{}}
\label{sec:boxqte}

Our exact quantifier identifies the set of inputs that violate the given properties.
We use $X_{\kappa}$ to denote the set of all inputs with prediction confidence exceeding $\kappa$:
\begin{equation}
\begin{aligned}
X_{\kappa} := \{\vec{x} \in X \mid f_c(\vec{x})>\kappa\}.
\end{aligned}
\end{equation}
Next, we denote $X_{\text{unfair}}$ as the set of all inputs with prediction confidence greater than $\kappa$, where there exists another input such that the input pair violates the $\epsilon$-$\kappa$-fairness:
\begin{equation}
\begin{aligned}
X_{\text{unfair}} :=\{\vec{x} \in X \mid & \exists \vec{x'} \in X, \bigwedge_{i \in  A \setminus S} |x_i - x'_i| \leq \epsilon_i \land \bigvee_{j \in S} x_j \neq x'_j \\ & \land f_c(\vec{x}) > \kappa \land f(\vec{x}) \neq f(\vec{x'}) \}.
\end{aligned}
\end{equation}
The $\epsilon$-$\kappa$-fairness measure can be then calculated as: 
\begin{equation}
\begin{aligned}
\mathcal{M}_{\text{fair}} = \frac{|X_\kappa \setminus X_{\text{unfair}}|}{|X_\kappa|} = 1 - \frac{|X_{\text{unfair}}|}{|X_\kappa|}.
\end{aligned}
\end{equation}

However, directly constructing $X_\kappa$ and $X_{\text{unfair}}$ is infeasible due to the vast input space.
We propose an efficient quantification method to address this challenge.
Our quantifier leverages the nature of tree ensembles: all inputs within the same path tuple receive the same prediction and confidence.
Thus, we focus only on path tuple pairs containing input pairs that violate properties.
By utilizing the boxes of these path tuples, we can accurately construct $X_\kappa$ and $X_{\text{unfair}}$ without exhaustively checking all inputs.

\begin{figure}
  \centering
  \includegraphics[width=.95\columnwidth]
  {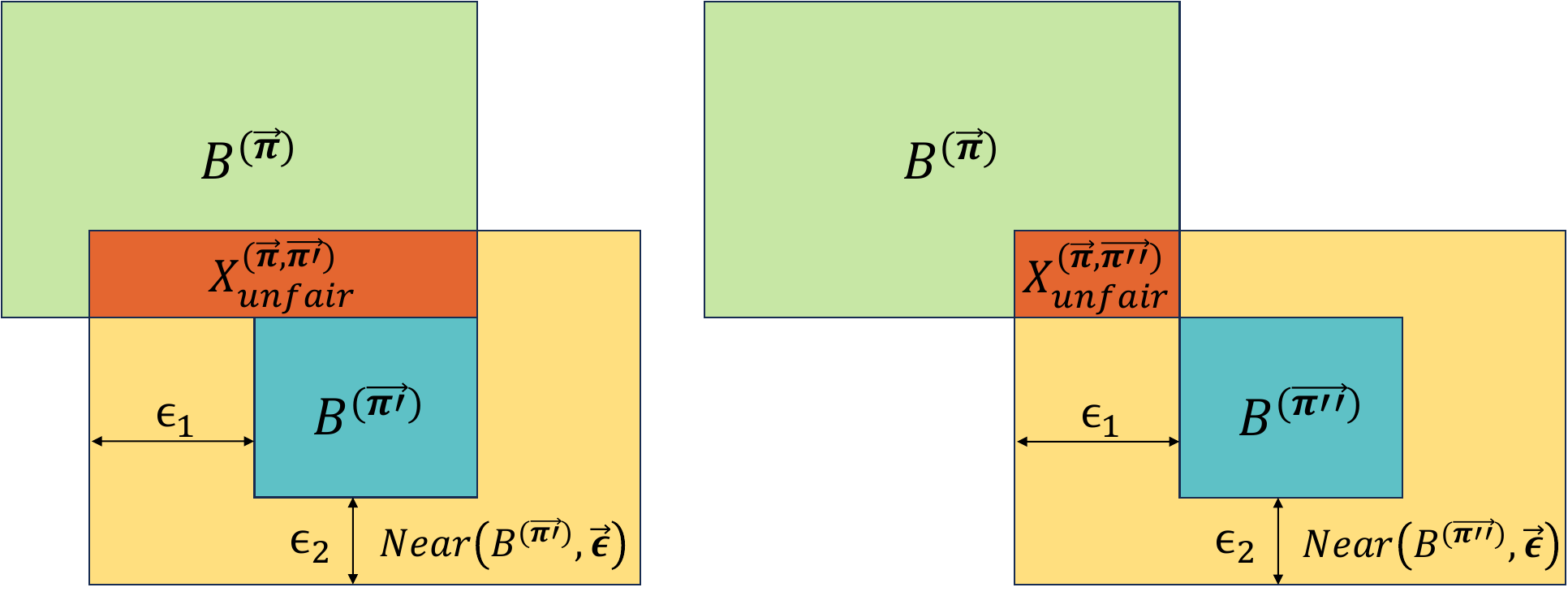}
  \caption{Two examples of partial counterexample regions (in red) obtained from box pairs.\label{fig:check_boxes}}
\end{figure}

We first define the \emph{$\vec{\epsilon}$-neighborhood} of a box $B^{(\vec{\pi})}$ along the $i$-th dimension as:
\begin{equation}
\begin{aligned}
\text{Near}(B_i^{(\vec{\pi})},\vec{\epsilon}) := [l_i^{(\vec{\pi})} - \epsilon_i, r_i^{(\vec{\pi})} + \epsilon_i).
\end{aligned}
\end{equation}
Based on the concept of $\vec{\epsilon}$-neighborhood, we introduce the following definition:

\begin{equation}
\begin{aligned}
X^{(\vec{\pi}, \vec{\pi'})}_{\text{unfair}} := \prod_{i=1}^n C_i, \text{ where }\label{X_fair}
\end{aligned}
\end{equation}
\begin{equation}
\begin{aligned}
C_i :=
\begin{cases} 
B_i^{(\vec{\pi})} \cap \text{Near}(B_i^{(\vec{\pi'})},\vec{\epsilon}), & \text{if } i\text{-th attribute is non-sensitive}; \\
B_i^{(\vec{\pi})} \setminus B_i^{(\vec{\pi'})}, & \text{if } i\text{-th attribute is sensitive, and } \\
 & \text{all sensitive attributes of } B^{{(\vec{\pi'})}} \\
 & \text{are restricted to a single value}; \\
B_i^{(\vec{\pi})}, & \text{otherwise},
\end{cases}
\end{aligned}
\end{equation}
where $B_i^{(\vec{\pi})}$ and $B_i^{(\vec{\pi'})}$ represent the intervals corresponding to the $i$-th dimension of the boxes $B^{(\vec{\pi})}$ and $B^{(\vec{\pi'})}$, respectively.
$X^{(\vec{\pi}, \vec{\pi'})}_{\text{unfair}}$ represents the set of all inputs in path tuple $\vec{\pi}$ that, when adjusted within the tolerance $\vec{\epsilon}$ for non-sensitive attributes and with at least one sensitive attribute modified, can fall into the path tuple $\vec{\pi'}$.
When $\vec{\pi}$ has a prediction confidence greater than $\kappa$ and $\vec{\pi}$ and $\vec{\pi'}$ receive different prediction outcomes, $X^{(\vec{\pi}, \vec{\pi'})}_{\text{unfair}}$ is a subset of $X_{\text{unfair}}$.
Hence, the entire $X_{\text{unfair}}$ can be constructed by enumerating every path tuple pair that meet the conditions.
\prettyref{fig:check_boxes} illustrates two examples of the regions obtained from two path tuple boxes along the non-sensitive attribute dimensions.

To construct $X_\kappa$, we introduce the following SMT formula, whose solutions correspond to inputs with prediction confidence greater than $\kappa$:
\begin{equation}
\begin{aligned}
\Phi_{\kappa}: = \Psi_{\text{trees}}(\vec{x}, \vec{v}, \vec{p}) \land \Psi_{\kappa}(\vec{v}, \kappa).\label{F_kappa}
\end{aligned}
\end{equation}
We also introduce the following SMT formula, whose solutions represent inputs with prediction confidence less than or equal to $\kappa$:
\begin{equation}
\begin{aligned}
\Phi_{\overline{\kappa}}: = \Psi_{\text{trees}}(\vec{x}, \vec{v}, \vec{p}) \land \neg \Psi_{\kappa}(\vec{v}, \kappa).\label{F_neg_kappa}
\end{aligned}
\end{equation}

$X_{\text{unfair}}$ can be fully constructed by enumerating all projected solutions of $\Phi_{\text{unfair}}$ on the variables $\vec{p}$ and $\vec{p'}$, and then utilizing the corresponding boxes and their $\vec{\epsilon}$-neighborhoods.
Similarly, $X_\kappa$ can be constructed by identifying all projected solutions of $\Phi_\kappa$ on the variables $\vec{p}$ and taking the union of their boxes.

\paragraph{Parallelization.}
While this method is more efficient than inspecting inputs, its performance is limited by the size of the tree ensembles.
As the tree ensemble grows, the number of potential path tuples increases exponentially.
To address this challenge, we accelerate the quantification process through parallelization.
Since the computation for each path tuple is completely independent, the inherent parallelism of the method is straightforward.

\paragraph{Any-time lower and upper bounds.}
The proposed method faces scalability challenges for large instances, as quantification itself is computationally difficult.
To mitigate this, we adapt the concept of \emph{any-time lower and upper bounds}, commonly employed in the literature (e.g., \cite{veritas+Laurens+ICML21}) for rapid estimation in computationally hard tasks.
Any-time bounds provide guaranteed lower and upper estimates of quantification at any point during computation.  

\SetKwProg{proc}{}{}{}
\begin{algorithm}[p]
\SetKw{And}{and} \SetKw{Or}{or} \SetKw{Not}{not} \SetKw{In}{in} \SetKw{Break}{break}
\caption{The algorithm of \boxmethod{}.\label{alg:quantification}}
\Parameter{tolerance $\vec{\epsilon}$ and confidence threshold $\kappa$.}
\KwData{a tree ensemble classifier $f$ with $m$ trees, its input space $X$ with $n$ attributes per input, and an index set of sensitive attributes $S$.}
\KwResult{quantitative measure $\mathcal{M}_{\text{fair}}$ for tree ensemble classifier $f$.}
\nonl \proc{$\Tool{Main}():$}{
$\mathcal{T}_X \gets |X|$; $\mathcal{T}_s \gets 0$; $\mathcal{T}_\kappa \gets 0$; $\mathcal{T}_{\overline{\kappa}} \gets 0$; $\text{LB} \gets 0$; $\text{UB} \gets 1$; \tcp{these variables are shared among tasks.}
%$\mathcal{T}_X \gets |X|$; $\mathcal{T}_s, \mathcal{T}_\kappa, \mathcal{T}_{\overline{\kappa}}, \text{LB}, \text{UB} \gets 0$; \tcp{shared among tasks.}
run $\Tool{Task}_\kappa()$ in parallel\;
run $\Tool{Task}_{\overline{\kappa}}()$ in parallel\;
\While{True}{
    \If{all tasks are finished}{\Return $\frac{\mathcal{T}_s}{\mathcal{T}_\kappa}$}
}
}
\nonl \proc{$\Tool{Task}_\kappa():$}{
$\Phi_\kappa \gets$ SMT formula encoded from Equation~\eqref{F_kappa}\;
$\mathcal{P} \gets \emptyset$\;
\While{$\Phi_\kappa$ is SAT}{
    $\vec{\pi} \gets$ a solution of $\Phi_\kappa$ projected onto variables $\vec{p}$\;
    $\mathcal{P} \gets \mathcal{P} \cup \{\vec{\pi}\}$\;
    $\Phi_\kappa \gets \Phi_\kappa \land \bigvee_{i=1}^m{p_i\neq \pi_i}$\;
    run $\Tool{Task}_{s}(\vec{\pi})$ in parallel\;
}
%$\mathcal{T}_{\overline{\kappa}} \gets 1 - \sum_{\vec{\pi} \in \mathcal{P}}\left| B^{(\vec{\pi})} \right|$\;
$\mathcal{T}_{\overline{\kappa}} \gets \mathcal{T}_X - \sum_{\vec{\pi} \in \mathcal{P}}\left| B^{(\vec{\pi})} \right|$\;
$LB \gets \frac{\mathcal{T}_\kappa - \mathcal{T}_s}{\mathcal{T}_X - \mathcal{T}_{\overline{\kappa}}}$; $UB \gets \frac{\mathcal{T}_X - \mathcal{T}_{\overline{\kappa}} - \mathcal{T}_s}{\mathcal{T}_X - \mathcal{T}_{\overline{\kappa}}}$\;
terminate all $\Tool{Task}_{\overline{\kappa}}$ tasks\;
}
\nonl \proc{$\Tool{Task}_s(\vec{\pi}):$}{
$\Phi_{\text{unfair}} \gets$ SMT formula encoded from Equation~\eqref{F_fair}\;
$\Phi_{\text{unfair}}^{(\vec{\pi})} \gets \Phi_{\text{unfair}} \land \bigwedge_{i=1}^m{p_i=\pi_i}$\;
$\mathcal{P'} \gets \emptyset$\;
\While{$\Phi_{\text{unfair}}^{(\vec{\pi})}$ is SAT}{
    $\vec{\pi'} \gets$ a solution of $\Phi_{\text{unfair}}^{(\vec{\pi})}$ projected onto variables $\vec{p'}$\;
    $\mathcal{P'} \gets \mathcal{P'} \cup \{\vec{\pi'}\}$\;
    $\Phi_{\text{unfair}}^{(\vec{\pi})} \gets \Phi_{\text{unfair}}^{(\vec{\pi})}\land \bigvee_{i=1}^m{p'_i\neq \pi'_i}$\label{line:replace-box-blocking}\;
}
$\mathcal{T}_{s} \gets \mathcal{T}_{s} + \left| \bigcup_{\vec{\pi'} \in \mathcal{P'}}X_{\text{unfair}}^{(\vec{\pi},\vec{\pi'})} \right|$\; 
%\tcp{$X_{\text{unfair}}^{(\vec{\pi},\vec{\pi'})}$ Equation~\eqref{X_robust} (for $\mathcal{M}_{\text{robust}}$) or \eqref{X_fair} (for $\mathcal{M}_{\text{fair}}$).}
$\mathcal{T}_{\kappa} \gets \mathcal{T}_{\kappa} + \left| B^{(\vec{\pi})} \right|$\;
$LB \gets \frac{\mathcal{T}_\kappa - \mathcal{T}_s}{\mathcal{T}_X - \mathcal{T}_{\overline{\kappa}}}$;
$UB \gets \frac{\mathcal{T}_X - \mathcal{T}_{\overline{\kappa}} - \mathcal{T}_s}{\mathcal{T}_X - \mathcal{T}_{\overline{\kappa}}}$\;
}
\nonl \proc{$\Tool{Task}_{\overline{\kappa}}():$}{
$\Phi_{\overline{\kappa}} \gets$ SMT formula encoded from Equation~\eqref{F_neg_kappa}\;
%$\mathcal{Q} \gets \emptyset$\;
\While{$\Phi_{\overline{\kappa}}$ is SAT}{
    $\vec{\pi} \gets$ a solution of $\Phi_{\overline{\kappa}}$ projected onto variables $\vec{p}$\;
    %$\mathcal{Q} \gets \mathcal{Q} \cup \{\vec{\pi}\}$\;
    $\Phi_{\overline{\kappa}} \gets \Phi_{\overline{\kappa}} \land \bigvee_{i=1}^m{p_i\neq \pi_i}$\;
    $\mathcal{T}_{\overline{\kappa}} \gets \mathcal{T}_{\overline{\kappa}} + \left| B^{(\vec{\pi})} \right|$\;
    $LB \gets \frac{\mathcal{T}_\kappa - \mathcal{T}_s}{\mathcal{T}_X - \mathcal{T}_{\overline{\kappa}}}$;
    $UB \gets \frac{\mathcal{T}_X - \mathcal{T}_{\overline{\kappa}} - \mathcal{T}_s}{\mathcal{T}_X - \mathcal{T}_{\overline{\kappa}}}$\;
}
}
\end{algorithm}

Given a path tuple $\vec{\pi}$ with confidence above $\kappa$, we define the SMT formula $\Phi_{\text{unfair}}^{(\vec{\pi})}$ as follows: 
\begin{equation}
\begin{aligned}
\Phi_{\text{unfair}}^{(\vec{\pi})} := \Phi_{\text{unfair}} \land \bigwedge_{i=1}^{m} p_i=\pi_i.\label{F_s_pi}
\end{aligned}
\end{equation}
This formula is used to identify all path tuples $\vec{\pi'}$ such that $\vec{\pi}$ and $\vec{\pi'}$ containing input pairs that violate properties.

\begin{theorem}[Any-Time Lower and Upper Bounds]\footnote{We introduce some useful notations to describe the solutions of SMT formulas.
Let $V$ be a set of variables, and let $F$ be an SMT formula defined over $V$.
We use $\allsol(F)$ to denote the set of all possible solutions to $F$. 
Given a subset of variables $W \subseteq V$, the projection of $\allsol(F)$ onto $W$ is denoted by $\allsol(F|W)$, which represents the set of solutions restricted to the variables in $W$.
For instance, $\allsol({\Phi_{\text{unfair}}| \vec{p}, \vec{p'})}$ represents the set of all possible solutions of $\Phi_{\text{unfair}}$ projected onto the path variables $\vec{p}$ and $\vec{p'}$. We use ${\vec{\pi}, \vec{\pi'} \in \allsol({\Phi_{\text{unfair}}| \vec{p}, \vec{p'})}}$ to represent that $(\vec{\pi}, \vec{\pi'})$ is one such projected solution.}
Given a tree ensemble classifier $f$ with input space $X$, 
let $\Phi_\kappa$, $\Phi_{\overline{\kappa}}$, and $\Phi_{\text{unfair}}^{(\vec{\pi})}$ be the SMT formula encoded from Equation~\eqref{F_kappa}, \eqref{F_neg_kappa}, and \eqref{F_s_pi}, respectively.
Then, for any $\mathcal{P} \subseteq \allsol(\Phi_\kappa|\vec{p})$ and any $\mathcal{Q} \subseteq \allsol(\Phi_{\overline{\kappa}}|\vec{p})$,
\begin{equation*}
\begin{aligned}
\displaystyle
\frac{\mathcal{T}_\kappa - \mathcal{T}_s}{\mathcal{T}_X - \mathcal{T}_{\overline{\kappa}}}
\leq
\mathcal{M}_{\text{fair}}
\leq
\frac{\mathcal{T}_X - \mathcal{T}_{\overline{\kappa}} - \mathcal{T}_s}{\mathcal{T}_X - \mathcal{T}_{\overline{\kappa}}},
\end{aligned}
\end{equation*}
where $\mathcal{T}_X = |X|$,
$\mathcal{T}_s = \sum_{\vec{\pi}\in \mathcal{P}} \left|\bigcup_{\vec{\pi'} \in \allsol({\Phi_{\text{unfair}}^{(\vec{\pi})}| \vec{p'}})} X_{\text{unfair}}^{(\vec{\pi}, \vec{\pi'})}\right|$, 
$\mathcal{T}_\kappa = \sum_{\vec{\pi}\in \mathcal{P}} \left| B^{(\vec{\pi})} \right|$, 
and $\mathcal{T}_{\overline{\kappa}} = \sum_{\vec{\pi}\in \mathcal{Q}} \left| B^{(\vec{\pi})} \right|$.\label{theo:bounds}
\end{theorem}
The proof of Theorem~\ref{theo:bounds} is omitted.
Theorem~\ref{theo:bounds} establishes a method for computing the upper and lower bounds of the measure $\mathcal{M}_{\text{fair}}$ at any time during the solving process.
Notably, in addition to solving $\Phi_\kappa$ and $\Phi_{\text{unfair}}^{(\vec{\pi})}$, the theorem also demonstrates the utility of solving $\Phi_{\overline{\kappa}}$,  as its value further contributes to refining the bounds.
\prettyref{alg:quantification} provides a detailed description of our exact quantification method. 
The algorithm includes three types of tasks: $\Tool{Task}_\kappa$, which solves $\Phi_\kappa$, $\Tool{Task}_s$, which solves $\Phi_{\text{unfair}}^{(\vec{\pi})}$, and $\Tool{Task}_{\overline{\kappa}}$, which solves $\Phi_{\overline{\kappa}}$.

\subsubsection{Enhancement techniques}

\paragraph{Task priority.}
We set a limit on the number of tasks that can be executed in parallel, requiring a queue to manage pending tasks. 
To accelerate the improvement of bound precision, we assign priority to each task $\Phi_{\text{unfair}}^{(\vec{\pi})}$, where its priority is determined by the size of the box of path tuple $\vec{\pi}$, i.e., $|B^{(\vec{\pi})}|$.
This strategy is based on the following observation.
The difference between the upper and lower bounds, denoted as $\delta$, can be computed as:
\begin{equation}
    \delta = \frac{\mathcal{T}_X - \mathcal{T}_{\overline{\kappa}} - \mathcal{T}_s}{\mathcal{T}_X - \mathcal{T}_{\overline{\kappa}}} - \frac{\mathcal{T}_\kappa - \mathcal{T}_s}{\mathcal{T}_X - \mathcal{T}_{\overline{\kappa}}} = 1-\frac{\mathcal{T}_\kappa}{\mathcal{T}_X - \mathcal{T}_{\overline{\kappa}}}.
\end{equation}
From this, we observe that the more $\mathcal{T}_\kappa$ increases, the more $\delta$ shrinks.
Since $\mathcal{T}_\kappa$ updates only after completing an $\Phi_{\text{unfair}}^{(\vec{\pi})}$ task, and the update magnitude corresponds to the box size of $\vec{\pi}$, i.e., $\mathcal{T}_\kappa = \mathcal{T}_\kappa + |B^{(\vec{\pi})}|$, prioritizing $\Phi_{\text{unfair}}^{(\vec{\pi})}$ with larger boxes is expected to accelerate the growth of $\mathcal{T}_\kappa$, leading to a faster reduction in $\delta$.

\paragraph{Task decomposition.}
In the \prettyref{alg:quantification}, all three types of tasks aim to enumerate all solutions of the SMT formula.
We adopt the strategy of incrementally adding blocking constraints to rule out solutions that have already been identified. 
However, the increasing number of blocking constraints may degrade the solving performance.
To address this issue, we impose an upper limit on the number of blocking constraints for each task.
Once this limit is exceeded, the task is divided into sub-tasks with non-overlapping solution spaces, and the blocking constraints are accordingly allocated among these sub-tasks.
Furthermore, these sub-tasks can be further divided if their blocking constraints also exceed the specified limit.

\paragraph{Box blcoking.}

Redundant box checking can be avoided.
In \prettyref{fig:check_boxes}, for example, once the two left-hand boxes are examined, the right-hand pair can be omitted since $X^{(\vec{\pi}, \vec{\pi''})}_\text{unfair} \subseteq X^{(\vec{\pi}, \vec{\pi'})}_\text{unfair}$.
After checking $B^{(\vec{\pi})}$ and $B^{(\vec{\pi'})}$,
the fowllowing constraint, called \emph{box blocking}, can rule out these redundant checks:
\begin{equation}
\Phi_{\text{outbox}}^{(\vec{\pi'})} := \bigvee_{i=1}^n{(x_i < l_i^{(\vec{\pi'})} - \epsilon_i) \land  (x_i \geq r_i^{(\vec{\pi'})} + \epsilon_i)}.
\end{equation}
We implement box blocking by replacing \prettyref{line:replace-box-blocking} of \prettyref{alg:quantification} with the fowllowing equation:
\begin{equation}
\begin{aligned}
\Phi_{\text{unfair}}^{(\vec{\pi})} \gets 
\Phi_{\text{unfair}}^{(\vec{\pi})} \land 
\bigvee_{i=1}^m{p'_i\neq \pi'_i} \land
\Phi_{\text{outbox}}^{(\vec{\pi')}}.
\end{aligned}
\end{equation}

\section{Experimental Evaluation\label{sec:experiment}}

\newcommand{\RQone}{How does \boxmethod{} perform compared to state-of-the-art quantification algorithms?}
\newcommand{\RQtwo}{What contributions do the three enhancement techniques make to the performance of \boxmethod{}?}
\newcommand{\RQthree}{How does \boxmethod{} perform in fairness testing compared to state-of-the-art fairness testing algorithms?}

This section evaluates our proposed evaluation algorithm, \boxmethod{}, by addressing the fowllowing three research questions (RQs).
\begin{description}
    \item[\textbf{RQ1:}] \RQone{}
    \item[\textbf{RQ2:}] \RQtwo{}
    \item[\textbf{RQ3:}] \RQthree{}
\end{description}

\subsection{Experimental Setup}
\paragraph{Datasets and models}
Our experiments utilize the fowllowing five tabular datasets, which are widely used in fairness research.
\begin{description}[leftmargin=2em]
\item[$\bullet$ Census Income (Adult)~\cite{misc_adult_2}] is extracted from the 1994 U.S. Census database. It contains $32{,}561$ samples and $13$ attributes. The task is to predict whether an individual earns more than \$50K per year. The sensitive attributes incldue \textit{gender}, \textit{race} and \textit{age}.
\item[$\bullet$ Bank Marketing (Bank)~\cite{misc_bank}] comes from a Portuguese bank’s direct marketing campaigns with $45{,}211$ samples and $16$ attributes. The task is to predict whether a client will subscribe to a term deposit. The only sensitive attribute is \textit{age}.
\item[$\bullet$ COMPAS~\cite{misc_compas}] is a recidivism prediction dataset derived from a commercial risk assessment tool used in the U.S. criminal justice system. After preprocessing, it contains approximately $6{,}907$ samples and $6$ attributes. The sensitive attributes are \textit{gender} and \textit{race}.
\item[$\bullet$ German Credit (Credit)~\cite{misc_german}] contains $600$ samples with $20$ attributes, used to classify individuals as having good or bad credit risk. The sensitive attributes are \textit{gender} and \textit{age}.
\item[$\bullet$ LSAC~\cite{misc_lsac}] is based on the National Longitudinal Bar Passage Study conducted by the Law School Admission Council from 1991 to 1997, tracking academic and demographic information of law students. The dataset contains $26{,}551$ samples and $11$ attributes. The sensitive attributes include \textit{gender} and \textit{race}.
\end{description}
Although our method can be applied to various tree ensembles, we evaluate it using GBDTs due to their widespread adoption and strong empirical performance.
We train XGBoost framework~\cite{chen+KDD16} to train GBDTs on the above five datasets.
Regarding hyperparameters, the number of trees is set to $50$ and the maximum depth is set to $3$. We have confirmed that further increasing these parameters does not lead to significant improvements in accuracy.

\SetKwProg{proc}{}{}{}
\begin{algorithm}[t]
\SetKw{And}{and} \SetKw{Or}{or} \SetKw{Not}{not} \SetKw{In}{in} \SetKw{Break}{break}
\caption{The algorithm of \cegarmethod{}.}
\label{alg:cegar}
\KwData{a tree ensemble classifier $f$ and its input space $X$.}
\KwResult{$\text{LB}$, Lower bound of quantitative measure $\mathcal{M}_{\text{fair}}$ for tree ensemble classifier $f$.}
{
$\mathcal{T}_X \gets |X|$; $\mathcal{T}_{fair} \gets 0$; $\text{LB} \gets \frac{\mathcal{T}_{fair}}{\mathcal{T}_X}$\;
$Q \gets \text{empty queue}$\;
$Q.enqueue(X)$\;
\While{not timeout}{
    $R \gets Q.dequeue()$\;
    $F_{\text{space}} \gets \text{formula ensuring inputs are within } R$\;
    \If{$F_{\text{unfair}} \land F_{\text{space}}$ is SAT}{
        $R_l, R_r \gets refine(R)$\;
        $Q.enqueue(R_l)$; $Q.enqueue(R_r)$\;
    }\Else{
        $\mathcal{T}_{fair} \gets \mathcal{T}_{fair} + \text{size}(R)$; $\text{LB} \gets \frac{\mathcal{T}_{fair}}{\mathcal{T}_X}$\;
    }
}
\Return $\text{LB}$\;
}
\end{algorithm}

\paragraph{Baselines for RQ1}
To the best of our knowledge, there is currently no quantification algorithm specifically designed for tree ensembles.
However, methods such as DeepGemini~\cite{DeepGemini+AAAI23} and FairQuant~\cite{fairquant+ICSE25} have been proposed for quantifying fairness of DNNs.
The core iead of DeepGemini and FairQuant are based on the Counterexample-Guided Abstraction Refinement (CEGAR), which is a generic flamework that can be applied for other type of ML models.
We generalize their core idea and describe a CEGAR-based quantification algorithm called \cegarmethod{}.
\cegarmethod{} reduces the quantification problem to a series of verification tasks that determine whether counterexamples exist within specific regions of the input space.
Starting from the entire input space, if a region contains counterexamples, it is refined into two sub-regions for further analysis. If no counterexamples are found, the region is marked as a fair region. 
As the process proceeds, more fair regions are identified, which can be used to calculate lower bound of fairness measure.
The algorithm is presented in \prettyref{alg:cegar}.
As long as a way toverify whether a region contains counterexamples is accessible, \cegarmethod{} is model-agnostic. 
Clearly, \cegarmethod{} can also be used to our tree ensemble setting by using SMT encoding in section 3. Therefore, we adopt \cegarmethod{} as the state-of-the-art quantification baseline to address RQ1.
We prepare the fowllowing three variants of \cegarmethod{}, differing in their refinement strategies.
\begin{description}[leftmargin=2em]
\item[$\bullet$ \cegarmethod{ (random)}.]
This variant adopts \emph{random refinement}, as proposed by \cite{DeepGemini+AAAI23}. It randomly selects an non-sensitive attribute and evenly splits the region along that dimension.
\item[$\bullet$ \cegarmethod{ (node)}.]
Tailored for tree ensembles, this variant uses \emph{node refinement}, randomly selecting a node from the tree ensemble and partitioning the region according to its split condition. It may terminate early if the region is too small to allow any valid node-based split.
\item[$\bullet$ \cegarmethod{ (node+random)}.]
A hybrid strategy that first applies node refinement switches to random refinement when no further node-based splits are feasible.
\end{description}
Note that \cegarmethod{} naturally supports parallelization. For a fair comparison, we use its parallel version with the same maximum number of concurrent tasks as \boxmethod{}.

\paragraph{Baselines for RQ2}

To answer RQ2, we prepare three modified versions of \boxmethod{}: \textbf{\boxmethod{ (no-prior)}}, \textbf{\boxmethod{ (no-decomp)}}, and \textbf{\boxmethod{ (no-box-block)}},
Each variant omits one of the proposed enhancements: task prioritization, task decomposition, and box blocking, respectively.
The performance differences between these variants and the full \boxmethod{} indicate the contribution of each enhancement techniques.

\paragraph{Baselines for RQ3}
To the best of our knowledge, no white-box fairness testing technique exists for tree ensembles.
We therefore adopt several state-of-the-art black-box methods as baselines: 
\textbf{\themis{}}~\cite{Galhotra+FSE17}, \textbf{\sg{}}~\cite{Aggarwal+FSE19}, \textbf{\expga{}}~\cite{Fan+ICSE22}, and \textbf{\aft{}}~\cite{zhao+ASE24}.
Among them, \aft{} is reported as the most efficient method in its original paper.

\paragraph{Parameter settings for \boxmethod{} and baselines}
Both \boxmethod{} and \cegarmethod{} are executed in a parallel environment with a maximum of $17$ concurrent tasks. For \boxmethod{}, the resource allocation is as follows: one task for task management, two tasks for $\Tool{Task}_{\overline{\kappa}}$, four tasks for $\Tool{Task}_{\kappa}$, and ten tasks for $\Tool{Task}_{s}$.
The maximum blocking threshold for triggering task decomposition in \boxmethod{} is set to $100$.
Parameter settings for all fairness testing baselines follow the defaults reported in their original papers or public implementations.
For runtime configurations, quantification methods are given a timeout of 600s, while fairness testing techniques are allowed up to 3600s, which is a common practice in the domain.

\paragraph{Settings of $\vec{\epsilon}$ and $\kappa$}
This work focuses on tabular data, where each attribute varies in range and type.
For each attribute $i$, we set the tolerance $\epsilon_i$ to a fixed proportion of the domain range of that attribute, controlled by a parameter called epsilon rate ($\epsr{}$).
For categorical attributes, we set $\epsilon_i=0$, as tolerance is not meaningful for them.
Since some baselines do not support $\vec{\epsilon}$ or $\kappa$, we consider two configurations for each: $\epsr{}=0$ (i.e., no tolerance) and $\epsr{}=0.1$; and $\kappa=0.5$ (i.e., no confidence threshold) and $\kappa=0.7$. The specific settings vary depending on the RQs and baselines.

\paragraph{Experimental environment}
All experiments are conducted on a machine of Intel(R) Xeon(R) Silver 4210 CPU @ 2.20 GHz Processor, 20 CPUs, 32 GB memory, Ubuntu 20.04.6 LTS with Python 3.8.10. 
SMT solving was performed using Z3 \cite{Moura+08z3}.

\subsection{Results}
\label{sec:exp-results}

\begin{table}[t]
\caption{Quantification performance of \boxmethod{} and \cegarmethod{} with $\kappa=0.5$ and time limit of 600s.
`Conv.' indicates convergence (\cmark{}) or not (\xmark{}). `Gap' is the final width of the bounds interval: $0$ if converged, otherwise UB minus LB.}
\label{tab:res-box-vs-cegar}
\centering
\setlength{\tabcolsep}{4pt}
\renewcommand{\arraystretch}{1.2}
\resizebox{1\columnwidth}{!}{
\begin{tabular}{c|c|c||crr|crr|crr|crr}
\hline 
\multirow{3}{*}{Prop.} & \multirow{3}{*}{Dataset } & \multirow{3}{*}{Attr. } & \multicolumn{3}{c|}{\boxmethod{} } & \multicolumn{3}{c|}{\cegarmethod{} } & \multicolumn{3}{c|}{\cegarmethod{} } & \multicolumn{3}{c}{\cegarmethod{} }\tabularnewline
 &  &  & \multicolumn{3}{c|}{} & \multicolumn{3}{c|}{(random)} & \multicolumn{3}{c|}{(node)} & \multicolumn{3}{c}{(node+random)}\tabularnewline
\cline{4-15} \cline{5-15} \cline{6-15} \cline{7-15} \cline{8-15} \cline{9-15} \cline{10-15} \cline{11-15} \cline{12-15} \cline{13-15} \cline{14-15} \cline{15-15} 
 &  &  & Conv. & Gap  & \multicolumn{1}{c|}{Time} & Conv. & Gap  & Time  & Conv. & Gap  & Time  & Conv. & Gap  & Time\tabularnewline
\hline 
\multirow{10}{*}{\begin{turn}{90}
Fairness ($\epsr{}=0$ )
\end{turn}} & \multirow{3}{*}{Adult } & age  & \cmark{}  & \multicolumn{1}{r}{0.00} & 26.4 & \xmark{}  & 0.39 & 600.0 & \cmark{}  & 0.00 & 49.8 & \xmark{}  & <.01 & 600.0\tabularnewline
 &  & race  & \cmark{}  & \multicolumn{1}{r}{0.00} & <0.1 & \cmark{}  & 0.00 & <0.1 & \cmark{}  & 0.00 & <0.1 & \cmark{}  & 0.00 & <0.1\tabularnewline
 &  & gender  & \cmark{}  & \multicolumn{1}{r}{0.00} & <0.1 & \cmark{}  & 0.00 & <0.1 & \cmark{}  & 0.00 & <0.1 & \cmark{}  & 0.00 & <0.1\tabularnewline
\cline{2-15} \cline{3-15} \cline{4-15} \cline{5-15} \cline{6-15} \cline{7-15} \cline{8-15} \cline{9-15} \cline{10-15} \cline{11-15} \cline{12-15} \cline{13-15} \cline{14-15} \cline{15-15} 
 & Bank  & age  & \cmark{}  & \multicolumn{1}{r}{0.00} & 101.9 & \xmark{}  & 0.64 & 600.0 & \cmark{}  & 0.00 & 63.2 & \xmark{}  & 0.08 & 600.0\tabularnewline
\cline{2-15} \cline{3-15} \cline{4-15} \cline{5-15} \cline{6-15} \cline{7-15} \cline{8-15} \cline{9-15} \cline{10-15} \cline{11-15} \cline{12-15} \cline{13-15} \cline{14-15} \cline{15-15} 
 & \multirow{2}{*}{COMPAS} & race  & \cmark{}  & \multicolumn{1}{r}{0.00} & 34.0 & \cmark{}  & 0.00 & 251.6 & \cmark{}  & 0.00 & 11.3 & \cmark{}  & 0.00 & 314.0\tabularnewline
 &  & gender  & \cmark{}  & \multicolumn{1}{r}{0.00} & 33.5 & \cmark{}  & 0.00 & 126.0 & \cmark{}  & 0.00 & 5.1 & \cmark{}  & 0.00 & 158.0\tabularnewline
\cline{2-15} \cline{3-15} \cline{4-15} \cline{5-15} \cline{6-15} \cline{7-15} \cline{8-15} \cline{9-15} \cline{10-15} \cline{11-15} \cline{12-15} \cline{13-15} \cline{14-15} \cline{15-15} 
 & \multirow{2}{*}{Credit } & age  & \cmark{}  & \multicolumn{1}{r}{0.00} & 27.0 & \xmark{}  & 0.02 & 600.0 & \cmark{}  & 0.00 & 44.1 & \cmark{}  & 0.00 & 600.0\tabularnewline
 &  & gender  & \cmark{}  & \multicolumn{1}{r}{0.00} & 31.2 & \xmark{}  & 0.52 & 600.0 & \xmark{}  & 0.05 & 600.0 & \xmark{}  & 0.04 & 600.0\tabularnewline
\cline{2-15} \cline{3-15} \cline{4-15} \cline{5-15} \cline{6-15} \cline{7-15} \cline{8-15} \cline{9-15} \cline{10-15} \cline{11-15} \cline{12-15} \cline{13-15} \cline{14-15} \cline{15-15} 
 & \multirow{2}{*}{LSAC } & race  & \cmark{}  & \multicolumn{1}{r}{0.00} & 226.3 & \xmark{}  & 0.08 & 600.0 & \cmark{}  & 0.00 & 311.5 & \xmark{}  & <.01 & 600.0\tabularnewline
 &  & gender  & \cmark{}  & \multicolumn{1}{r}{0.00} & <0.1 & \cmark{}  & 0.00 & <0.1 & \cmark{}  & 0.00 & <0.1 & \cmark{}  & 0.00 & <0.1\tabularnewline
\hline 
\hline 
\multirow{10}{*}{\begin{turn}{90}
Fairness ($\epsr{}=0.1$)
\end{turn}} & \multirow{3}{*}{Adult } & age  & \cmark{}  & \multicolumn{1}{r}{0.00} & 121.9 & \xmark{}  & 0.52 & 600.0 & \xmark{}  & 1.00 & 192.8 & \xmark{}  & 0.97 & 600.0\tabularnewline
 &  & race  & \cmark{}  & \multicolumn{1}{r}{0.00} & 89.7 & \xmark{}  & 0.62 & 600.0 & \xmark{}  & 0.99 & 192.4 & \xmark{}  & 0.93 & 600.0\tabularnewline
 &  & gender  & \cmark{}  & \multicolumn{1}{r}{0.00} & 89.7 & \xmark{}  & 0.67 & 600.0 & \xmark{}  & 0.99 & 192.6 & \xmark{}  & 0.96 & 600.0\tabularnewline
\cline{2-15} \cline{3-15} \cline{4-15} \cline{5-15} \cline{6-15} \cline{7-15} \cline{8-15} \cline{9-15} \cline{10-15} \cline{11-15} \cline{12-15} \cline{13-15} \cline{14-15} \cline{15-15} 
 & Bank  & age  & \cmark{}  & \multicolumn{1}{r}{0.00} & 187.6 & \xmark{}  & 0.71 & 600.0 & \xmark{}  & 0.22 & 100.2 & \xmark{}  & 0.23 & 600.0\tabularnewline
\cline{2-15} \cline{3-15} \cline{4-15} \cline{5-15} \cline{6-15} \cline{7-15} \cline{8-15} \cline{9-15} \cline{10-15} \cline{11-15} \cline{12-15} \cline{13-15} \cline{14-15} \cline{15-15} 
 & \multirow{2}{*}{COMPAS} & race  & \cmark{}  & \multicolumn{1}{r}{0.00} & 74.8 & \xmark{}  & 0.20 & 600.0 & \xmark{}  & 0.21 & 14.0 & \xmark{}  & 0.20 & 600.0\tabularnewline
 &  & gender  & \cmark{}  & \multicolumn{1}{r}{0.00} & 43.4 & \xmark{}  & 0.18 & 600.0 & \xmark{}  & 0.19 & 18.4 & \xmark{}  & 0.18 & 600.0\tabularnewline
\cline{2-15} \cline{3-15} \cline{4-15} \cline{5-15} \cline{6-15} \cline{7-15} \cline{8-15} \cline{9-15} \cline{10-15} \cline{11-15} \cline{12-15} \cline{13-15} \cline{14-15} \cline{15-15} 
 & \multirow{2}{*}{Credit } & age  & \cmark{}  & \multicolumn{1}{r}{0.00} & 87.0 & \xmark{} & 0.96 & 600.0 & \xmark{} & 0.75 & 600.0 & \xmark{} & 0.71 & 600.0\tabularnewline
 &  & gender  & \cmark{}  & \multicolumn{1}{r}{0.00} & 113.6 & \xmark{}  & 0.92 & 600.0 & \xmark{}  & 0.75 & 600.0 & \xmark{}  & 0.81 & 600.0\tabularnewline
\cline{2-15} \cline{3-15} \cline{4-15} \cline{5-15} \cline{6-15} \cline{7-15} \cline{8-15} \cline{9-15} \cline{10-15} \cline{11-15} \cline{12-15} \cline{13-15} \cline{14-15} \cline{15-15} 
 & \multirow{2}{*}{LSAC } & race  & \cmark{}  & \multicolumn{1}{r}{0.00} & 557.2 & \xmark{}  & 0.37 & 600.0 & \xmark{}  & 0.49 & 600.0 & \xmark{}  & 0.56 & 600.0\tabularnewline
 &  & gender  & \cmark{}  & \multicolumn{1}{r}{0.00} & 542.6 & \xmark{}  & 0.53 & 600.0 & \xmark{}  & 0.51 & 600.0 & \xmark{}  & 0.52 & 600.0\tabularnewline
\hline 
\hline 
\multirow{5}{*}{\begin{turn}{90}
Robust. ($\epsr{}=0.1$)
\end{turn}} & Adult  & ---  & \cmark{}  & 0.00 & 89.4 & \xmark{}  & 0.34 & 600.0 & \xmark{}  & 0.99 & 189.3 & \xmark{}  & 0.93 & 600.0\tabularnewline
\cline{2-15} \cline{3-15} \cline{4-15} \cline{5-15} \cline{6-15} \cline{7-15} \cline{8-15} \cline{9-15} \cline{10-15} \cline{11-15} \cline{12-15} \cline{13-15} \cline{14-15} \cline{15-15} 
 & Bank  & ---  & \cmark{}  & 0.00 & 137.3 & \xmark{}  & 0.64 & 600.0 & \xmark{}  & 0.22 & 93.7 & \xmark{}  & 0.22 & 600.0\tabularnewline
\cline{2-15} \cline{3-15} \cline{4-15} \cline{5-15} \cline{6-15} \cline{7-15} \cline{8-15} \cline{9-15} \cline{10-15} \cline{11-15} \cline{12-15} \cline{13-15} \cline{14-15} \cline{15-15} 
 & COMPAS  & ---  & \cmark{}  & 0.00 & 39.8 & \xmark{}  & 0.18 & 600.0 & \xmark{}  & 0.19 & 16.4 & \xmark{}  & 0.19 & 600.0\tabularnewline
\cline{2-15} \cline{3-15} \cline{4-15} \cline{5-15} \cline{6-15} \cline{7-15} \cline{8-15} \cline{9-15} \cline{10-15} \cline{11-15} \cline{12-15} \cline{13-15} \cline{14-15} \cline{15-15} 
 & Credit  & ---  & \cmark{}  & 0.00 & 82.8 & \xmark{}  & 0.94 & 600.0 & \xmark{}  & 0.80 & 600.0 & \xmark{}  & 0.77 & 600.0\tabularnewline
\cline{2-15} \cline{3-15} \cline{4-15} \cline{5-15} \cline{6-15} \cline{7-15} \cline{8-15} \cline{9-15} \cline{10-15} \cline{11-15} \cline{12-15} \cline{13-15} \cline{14-15} \cline{15-15} 
 & LSAC  & ---  & \cmark{}  & 0.00 & 540.5 & \xmark{}  & 0.38 & 600.0 & \xmark{}  & 0.37 & 600.0 & \xmark{}  & 0.45 & 600.0\tabularnewline
\hline 
\hline 
\multicolumn{3}{c||}{\#Conv. / Avg. / Avg.} & 25 & \multicolumn{1}{r}{0.00} & 131.1 & 5 & 0.39 & 495.1 & 9 & 0.35 & 227.8 & 6 & 0.35 & 498.9\tabularnewline
\hline 
\end{tabular}
}
\end{table}

\begin{figure*}[t]
  \centering
    % line 1
    \begin{subfigure}[b]{0.24\textwidth}
        \centering
        \includegraphics[width=\textwidth]{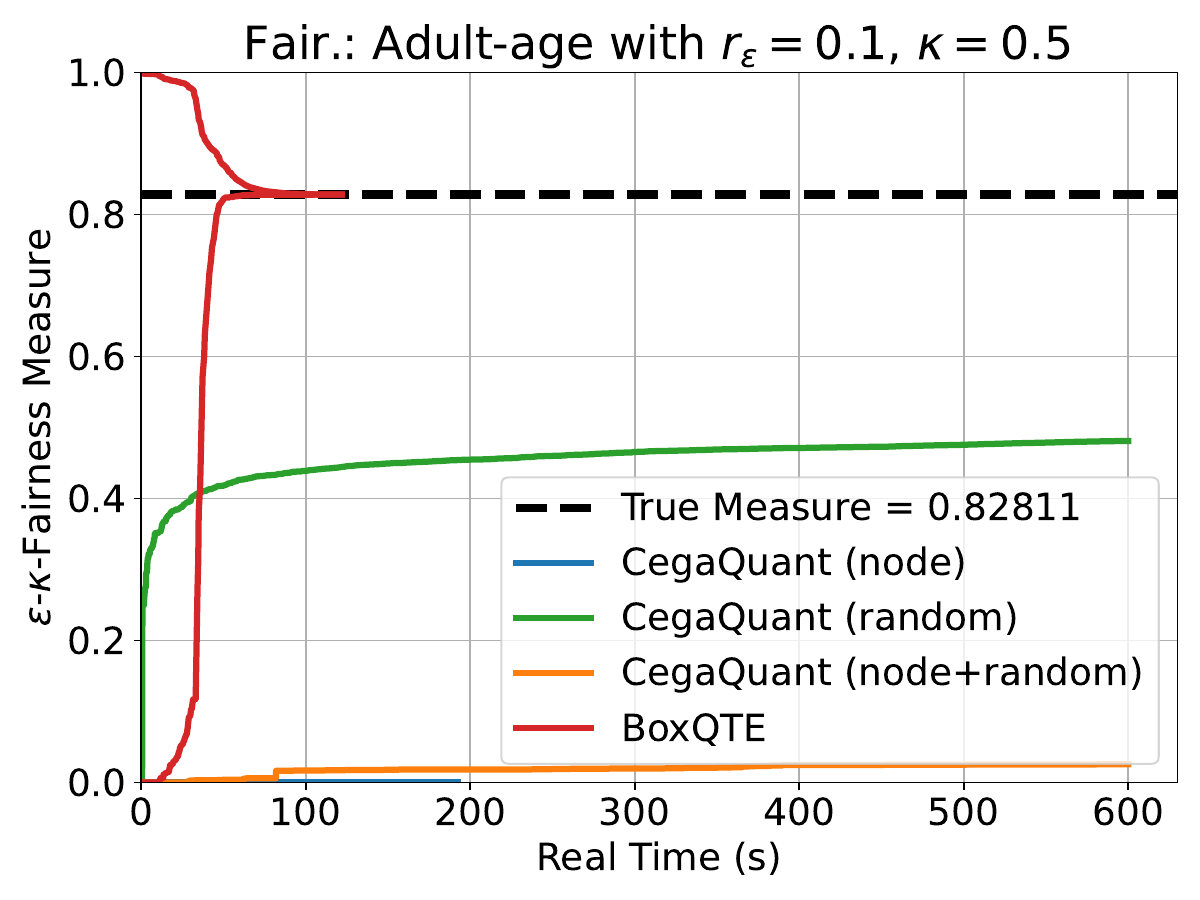}
        %\caption{1}
        %\label{fig:image2}
    \end{subfigure}
    \hfill
    \begin{subfigure}[b]{0.24\textwidth}
        \centering
        \includegraphics[width=\textwidth]{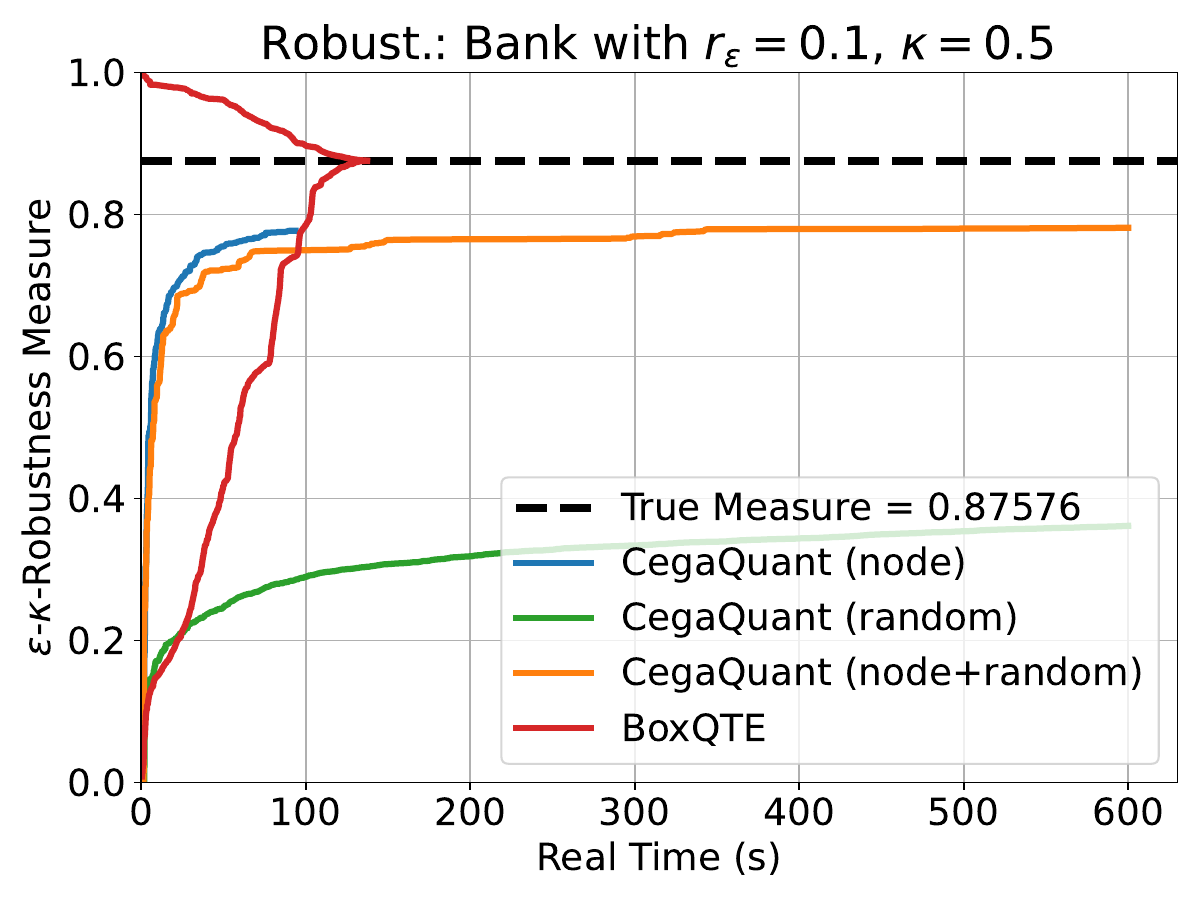}
        %\caption{1}
        %\label{fig:image2}
    \end{subfigure}
    \hfill
    \begin{subfigure}[b]{0.24\textwidth}
        \centering
        \includegraphics[width=\textwidth]{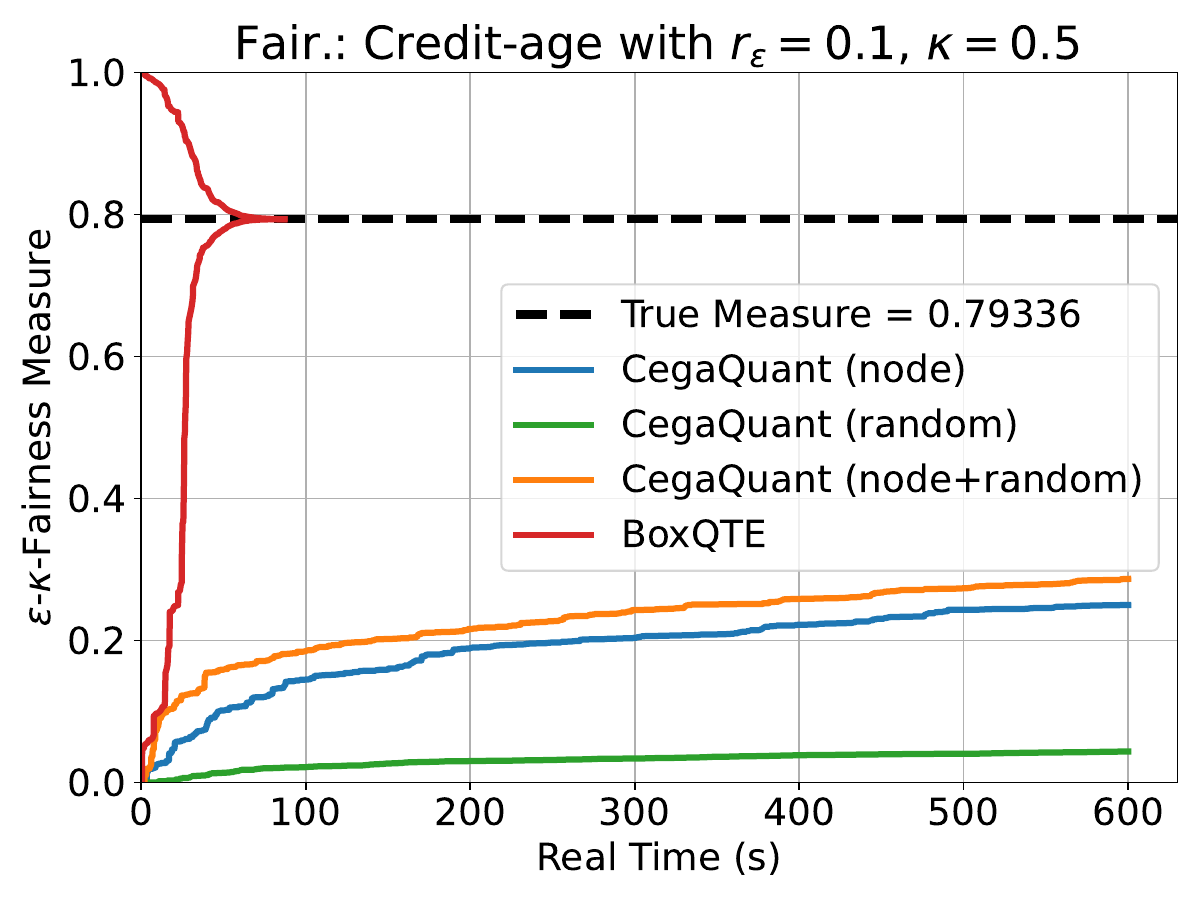}
        %\caption{1}
        %\label{fig:image2}
    \end{subfigure}
    \hfill
    \begin{subfigure}[b]{0.24\textwidth}
        \centering
        \includegraphics[width=\textwidth]{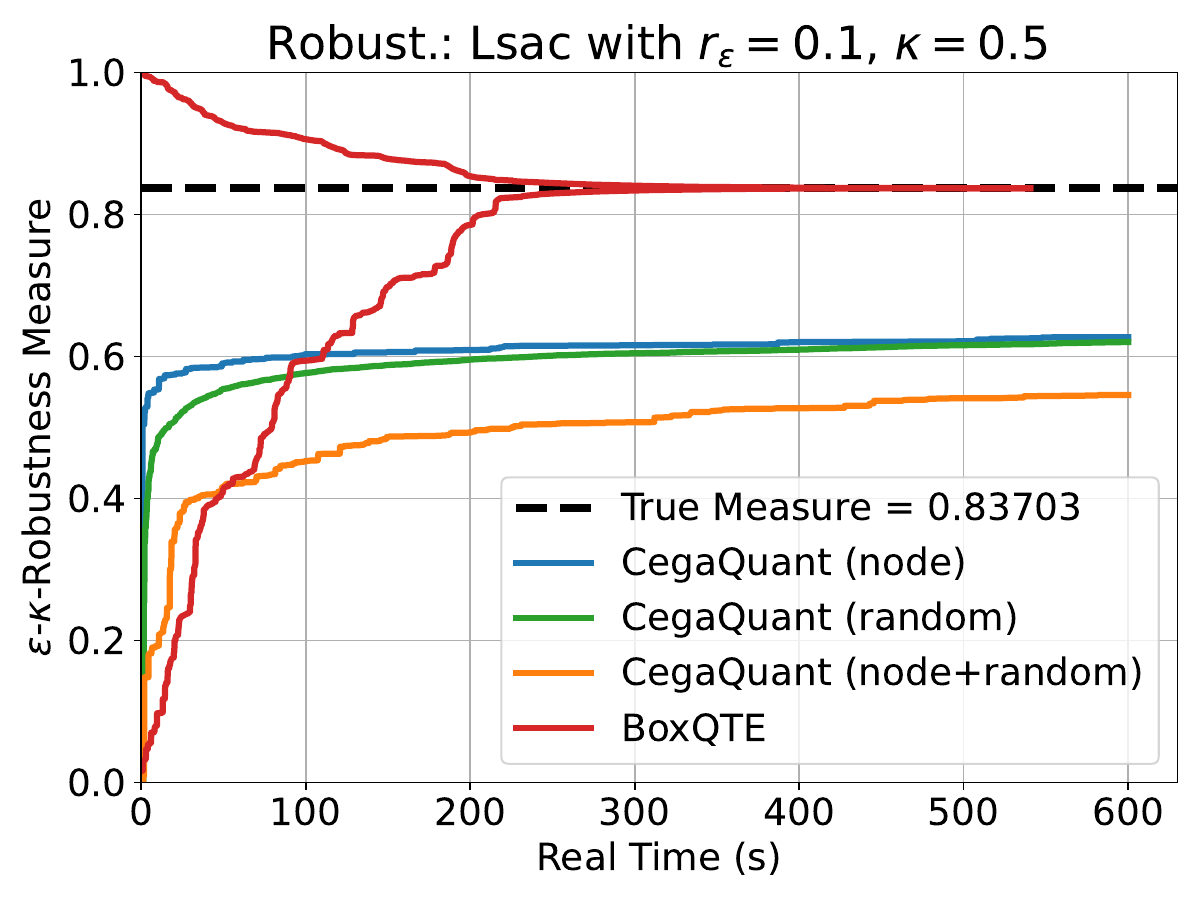}
        %\caption{1}
        %\label{fig:image2}
    \end{subfigure}
    % line 2
    \begin{subfigure}[b]{0.24\textwidth}
        \centering
        \includegraphics[width=\textwidth]{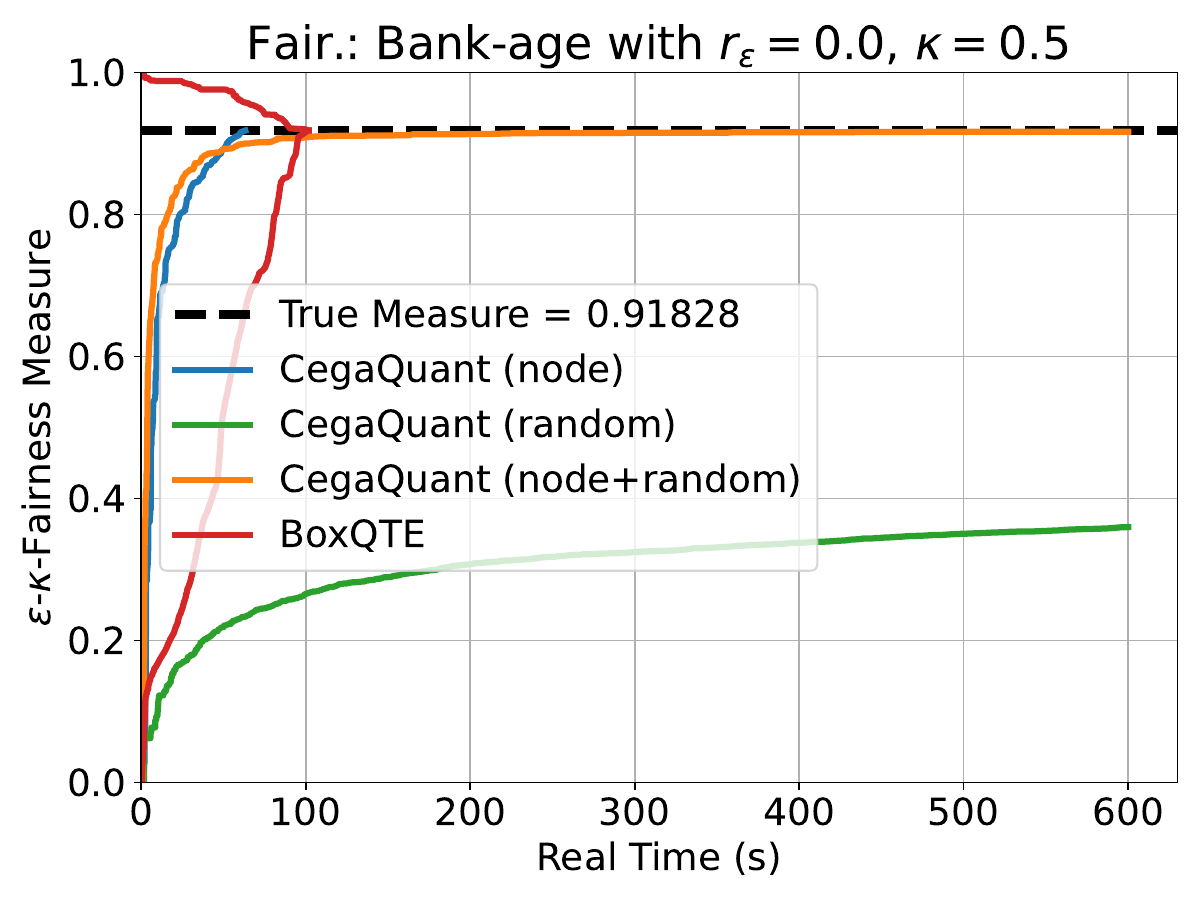}
        %\caption{1}
        %\label{fig:image2}
    \end{subfigure}
    \hfill
    \begin{subfigure}[b]{0.24\textwidth}
        \centering
        \includegraphics[width=\textwidth]{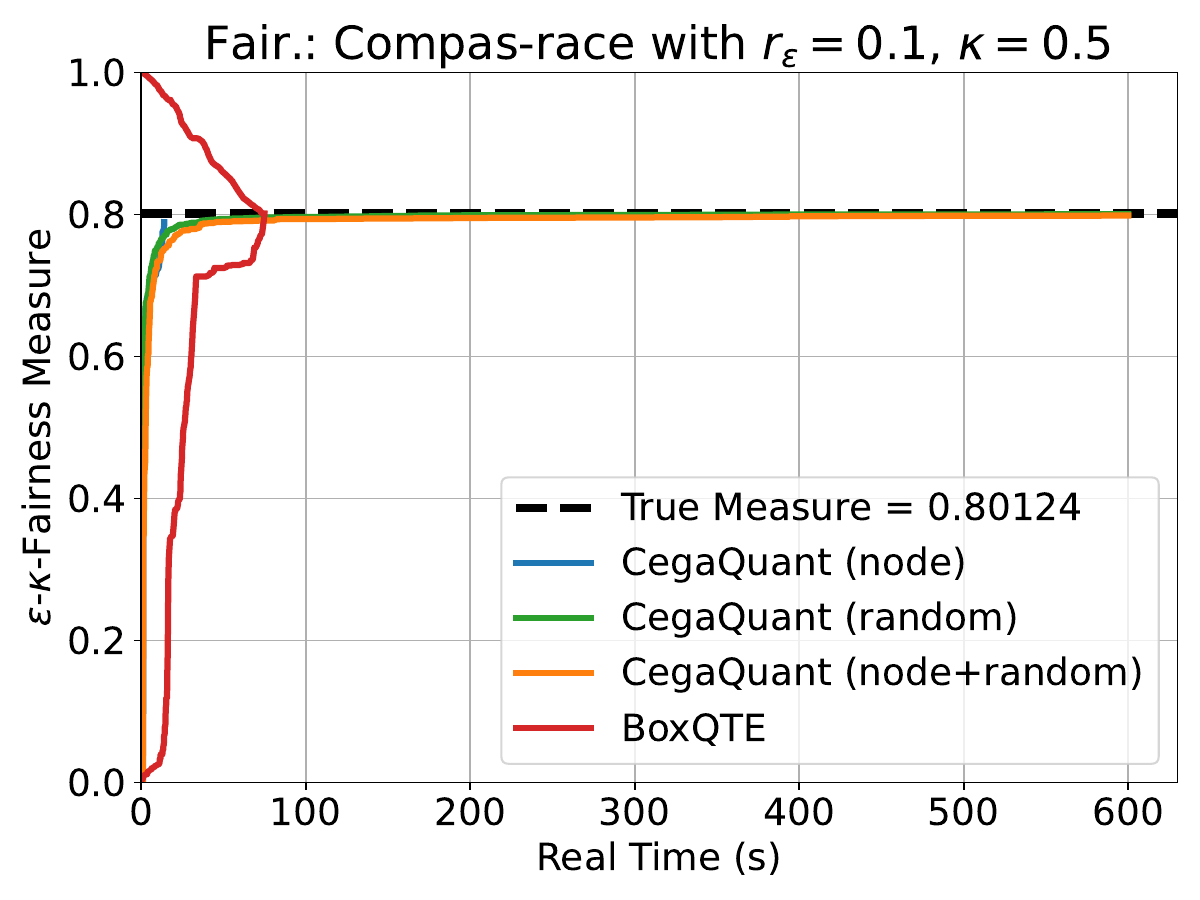}
        %\caption{1}
        %\label{fig:image2}
    \end{subfigure}
    \hfill
    \begin{subfigure}[b]{0.24\textwidth}
        \centering
        \includegraphics[width=\textwidth]{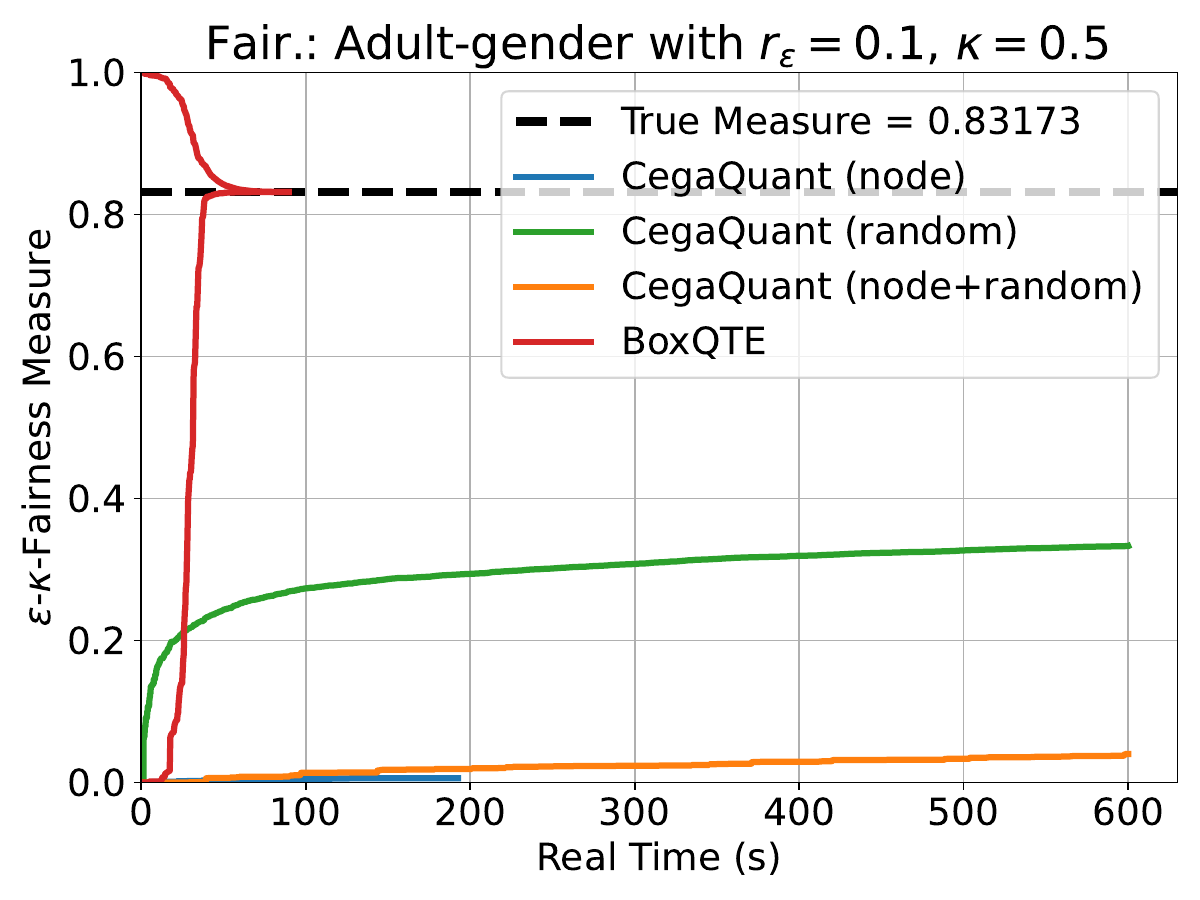}
        %\caption{1}
        %\label{fig:image2}
    \end{subfigure}
    \hfill
    \begin{subfigure}[b]{0.24\textwidth}
        \centering
        \includegraphics[width=\textwidth]{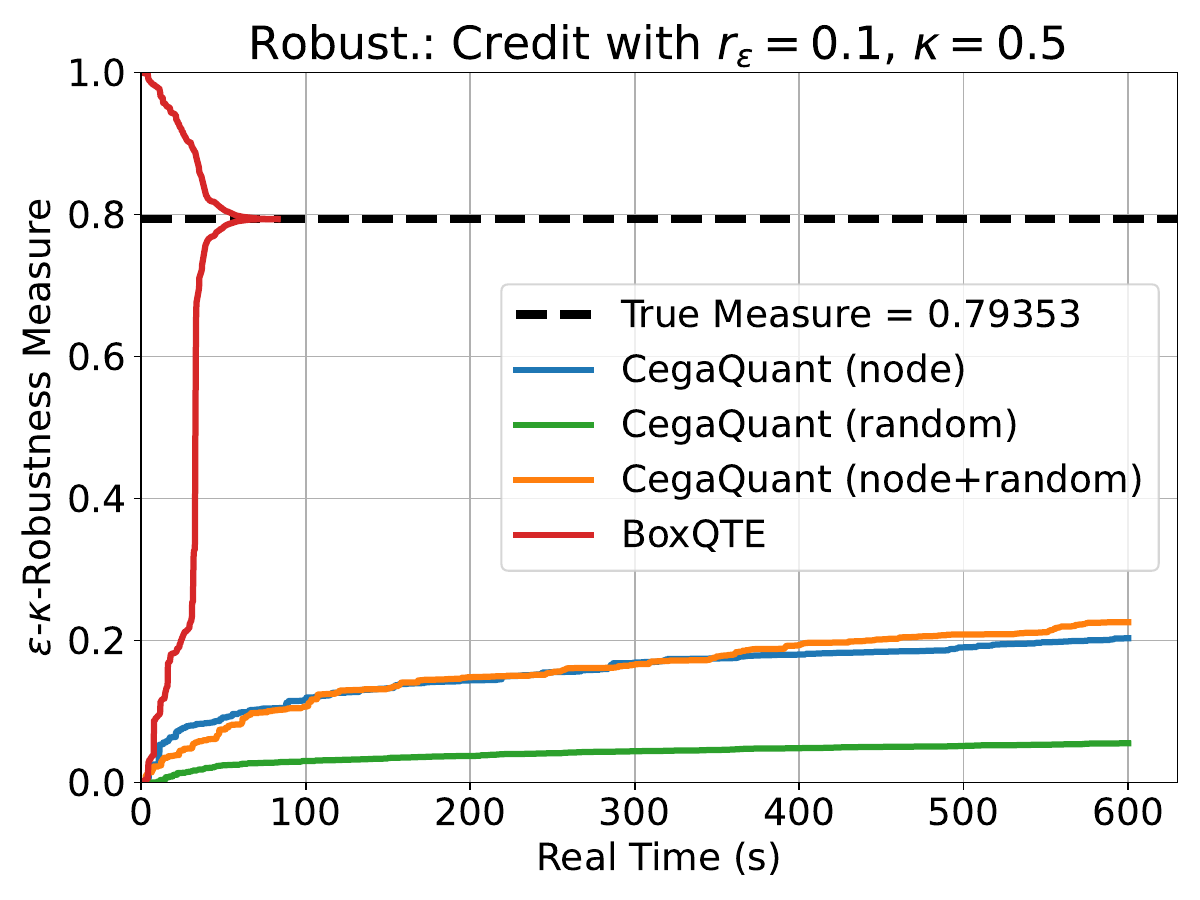}
        %\caption{1}
        %\label{fig:image2}
    \end{subfigure}
  \caption{Evolution of bounds over time of \boxmethod{} and \cegarmethod{} with time limit of 600s.}
  \label{fig:res-evolution-box-vs-cegar}
\end{figure*}

\subsubsection{RQ1: \RQone{}}
To answer RQ1, we conduct quantization using both our proposed method \boxmethod{} and the state-of-the-art quantitative method \cegarmethod{}, and compare their performance. 
Since \cegarmethod{} does not support handling the confidence threshold $\kappa$\footnote{In fact,  we have extended \cegarmethod{} to support $\kappa$, and experimental results demonstrate that even with $\kappa$ taken into account, \boxmethod{} still outperforms \cegarmethod{}, which is consistent with the conclusions of this work. Due to space limitations, the details of the extension and the experiments are not reported in this paper.}, we exclude the consideration of $\kappa$ by setting it to $0.5$. For tolerance $\vec{\epsilon}$, we consider $\epsr{} = 0$ and $\epsr{} = 0.1$.

\prettyref{tab:res-box-vs-cegar} summarizes the performance of \boxmethod{} and three \cegarmethod{} variants across $25$ configurations. \prettyref{fig:res-evolution-box-vs-cegar} shows partial results illustrating the evolution of quantified bounds over time.

As shown in \prettyref{tab:res-box-vs-cegar}, 
\boxmethod{} converges on all $25$ configurations, while the three \cegarmethod{} variants converge on only $5$, $9$, and $6$, respectively.
Their final bound widths (`Gap') are $0.39$, $0.35$, and $0.35$. \boxmethod{} is also the fastest, with an average runtime of 131.1s, compared to 495.1s, 227.8s, and 498.9s.
Notably, with $\epsr{}=0.1$, none of the \cegarmethod{} variants converge on any of the $15$ settings. This may be due to the increased complexity of counterexample distributions under tolerance. As shown in \prettyref{fig:check_boxes}, overlaps between boxes and neighborhoods of other boxes complicate refinement, making it less effective. In contrast, \boxmethod{} leverages the box structure to better handle such cases.

From \prettyref{fig:res-evolution-box-vs-cegar}, we observe that \cegarmethod{} methods initially yield rapid lower bound growth, but slow down significantly over time. This behavior may stem from repeated region refinement, which quickly reduces region sizes. As the lower bound increment is closely tied to region sizes, such shrinkage hinders further progress.

\begin{tcolorbox}
[size=small, arc=0pt, boxsep=5pt,left=2pt,right=2pt,top=0pt,bottom=0pt]
\textbf{Answer for RQ1:}
    Our proposed quantification method, \boxmethod{}, leverages the structural properties of the tree ensembles, demonstrating superior efficiency and effectiveness compared to state-of-the-art quantification methods.
\end{tcolorbox}

\begin{table}[t]
\caption{Quantification performance of \boxmethod{} and its three enhancements with $\epsr{}=0.1$, $\kappa=0.7$, and time limit of 600s.}
\label{tab:res_box_vs_improvement}
\centering
\setlength{\tabcolsep}{4pt}
\renewcommand{\arraystretch}{1.2}
\resizebox{1\columnwidth}{!}{
\begin{tabular}{c|c|c||ccr|crr|crr|crr}
\hline 
\multirow{3}{*}{Prop.} & \multirow{3}{*}{Dataset } & \multirow{3}{*}{Attr. } & \multicolumn{3}{c|}{\boxmethod{} } & \multicolumn{3}{c|}{\boxmethod{} } & \multicolumn{3}{c|}{\boxmethod{} } & \multicolumn{3}{c}{\boxmethod{} }\tabularnewline
 &  &  & \multicolumn{3}{c|}{} & \multicolumn{3}{c|}{(no-box-block)} & \multicolumn{3}{c|}{(no-decomp)} & \multicolumn{3}{c}{(no-prior)}\tabularnewline
\cline{4-15} \cline{5-15} \cline{6-15} \cline{7-15} \cline{8-15} \cline{9-15} \cline{10-15} \cline{11-15} \cline{12-15} \cline{13-15} \cline{14-15} \cline{15-15} 
 &  &  & Done & Gap & \multicolumn{1}{c|}{Time} & Done & Gap & Time & Done & Gap & Time & Done & Gap & Time\tabularnewline
\hline 
\multirow{10}{*}{\begin{turn}{90}
Fairness
\end{turn}} & \multirow{3}{*}{Adult } & age  & \cmark{}  & \multicolumn{1}{r}{0.00} & 56.7 & \cmark{}  & 0.00 & 203.0 & \cmark{}  & 0.00 & 59.9 & \cmark{}  & 0.00 & 63.1\tabularnewline
 &  & race  & \cmark{}  & \multicolumn{1}{r}{0.00} & 45.9 & \cmark{}  & 0.00 & 160.9 & \cmark{}  & 0.00 & 44.3 & \cmark{}  & 0.00 & 45.6\tabularnewline
 &  & gender  & \cmark{}  & \multicolumn{1}{r}{0.00} & 44.1 & \cmark{}  & 0.00 & 162.1 & \cmark{}  & 0.00 & 45.1 & \cmark{}  & 0.00 & 47.4\tabularnewline
\cline{2-15} \cline{3-15} \cline{4-15} \cline{5-15} \cline{6-15} \cline{7-15} \cline{8-15} \cline{9-15} \cline{10-15} \cline{11-15} \cline{12-15} \cline{13-15} \cline{14-15} \cline{15-15} 
 & Bank  & age  & \cmark{}  & \multicolumn{1}{r}{0.00} & 54.3 & \cmark{}  & 0.00 & 53.2 & \cmark{}  & 0.00 & 319.3 & \cmark{}  & 0.00 & 53.2\tabularnewline
\cline{2-15} \cline{3-15} \cline{4-15} \cline{5-15} \cline{6-15} \cline{7-15} \cline{8-15} \cline{9-15} \cline{10-15} \cline{11-15} \cline{12-15} \cline{13-15} \cline{14-15} \cline{15-15} 
 & \multirow{2}{*}{COMPAS} & age  & \cmark{}  & \multicolumn{1}{r}{0.00} & 28.7 & \cmark{}  & 0.00 & 28.6 & \cmark{}  & 0.00 & 128.1 & \cmark{}  & 0.00 & 28.2\tabularnewline
 &  & gender  & \cmark{}  & \multicolumn{1}{r}{0.00} & 26.9 & \cmark{}  & 0.00 & 27.5 & \cmark{}  & 0.00 & 129.3 & \cmark{}  & 0.00 & 26.1\tabularnewline
\cline{2-15} \cline{3-15} \cline{4-15} \cline{5-15} \cline{6-15} \cline{7-15} \cline{8-15} \cline{9-15} \cline{10-15} \cline{11-15} \cline{12-15} \cline{13-15} \cline{14-15} \cline{15-15} 
 & \multirow{2}{*}{Credit } & age  & \cmark{}  & \multicolumn{1}{r}{0.00} & 18.0 & \cmark{}  & 0.00 & 27.4 & \cmark{}  & 0.00 & 47.3 & \cmark{}  & 0.00 & 18.0\tabularnewline
 &  & gender  & \cmark{}  & \multicolumn{1}{r}{0.00} & 18.7 & \cmark{}  & 0.00 & 27.4 & \cmark{}  & 0.00 & 46.5 & \cmark{}  & 0.00 & 18.2\tabularnewline
\cline{2-15} \cline{3-15} \cline{4-15} \cline{5-15} \cline{6-15} \cline{7-15} \cline{8-15} \cline{9-15} \cline{10-15} \cline{11-15} \cline{12-15} \cline{13-15} \cline{14-15} \cline{15-15} 
 & \multirow{2}{*}{LSAC } & age  & \cmark{}  & \multicolumn{1}{r}{0.00} & 505.3 & \xmark{}  & 0.11 & 600.6 & \xmark{}  & 0.52 & 599.8 & \cmark{}  & 0.00 & 531.0\tabularnewline
 &  & gender  & \cmark{}  & \multicolumn{1}{r}{0.00} & 497.9 & \xmark{}  & 0.11 & 604.7 & \xmark{}  & 0.54 & 599.7 & \cmark{}  & 0.00 & 527.8\tabularnewline
\hline 
\hline 
\multirow{5}{*}{\begin{turn}{90}
Robustness
\end{turn}} & Adult  & --- & \cmark{}  & 0.00 & 45.6 & \cmark{}  & 0.00 & 161.1 & \cmark{}  & 0.00 & 44.4 & \cmark{}  & 0.00 & 46.7\tabularnewline
\cline{2-15} \cline{3-15} \cline{4-15} \cline{5-15} \cline{6-15} \cline{7-15} \cline{8-15} \cline{9-15} \cline{10-15} \cline{11-15} \cline{12-15} \cline{13-15} \cline{14-15} \cline{15-15} 
 & Bank  & --- & \cmark{}  & 0.00 & 53.6 & \cmark{}  & 0.00 & 49.8 & \cmark{}  & 0.00 & 320.0 & \cmark{}  & 0.00 & 52.3\tabularnewline
\cline{2-15} \cline{3-15} \cline{4-15} \cline{5-15} \cline{6-15} \cline{7-15} \cline{8-15} \cline{9-15} \cline{10-15} \cline{11-15} \cline{12-15} \cline{13-15} \cline{14-15} \cline{15-15} 
 & COMPAS & --- & \cmark{}  & 0.00 & 26.2 & \cmark{}  & 0.00 & 25.3 & \cmark{}  & 0.00 & 134.2 & \cmark{}  & 0.00 & 26.9\tabularnewline
\cline{2-15} \cline{3-15} \cline{4-15} \cline{5-15} \cline{6-15} \cline{7-15} \cline{8-15} \cline{9-15} \cline{10-15} \cline{11-15} \cline{12-15} \cline{13-15} \cline{14-15} \cline{15-15} 
 & Credit  & --- & \cmark{}  & 0.00 & 18.4 & \cmark{}  & 0.00 & 26.7 & \cmark{}  & 0.00 & 47.2 & \cmark{}  & 0.00 & 17.0\tabularnewline
\cline{2-15} \cline{3-15} \cline{4-15} \cline{5-15} \cline{6-15} \cline{7-15} \cline{8-15} \cline{9-15} \cline{10-15} \cline{11-15} \cline{12-15} \cline{13-15} \cline{14-15} \cline{15-15} 
 & LSAC  & --- & \cmark{}  & 0.00 & 501.4 & \xmark{}  & 0.11 & 606.9 & \xmark{}  & 0.51 & 599.7 & \cmark{}  & 0.00 & 528.3\tabularnewline
\hline 
\hline 
\multicolumn{3}{c||}{\#Done / Avg. } & 15 & \multicolumn{1}{r}{0.00} & 129.4 & 12 & 0.02 & 184.4 & 12 & 0.10 & 211.0 & 15 & 0.00 & 135.3\tabularnewline
\hline 
\end{tabular}
}
\end{table}

\begin{figure*}[t]
  \centering
    \begin{subfigure}[b]{0.24\textwidth}
        \centering
        \includegraphics[width=\textwidth]{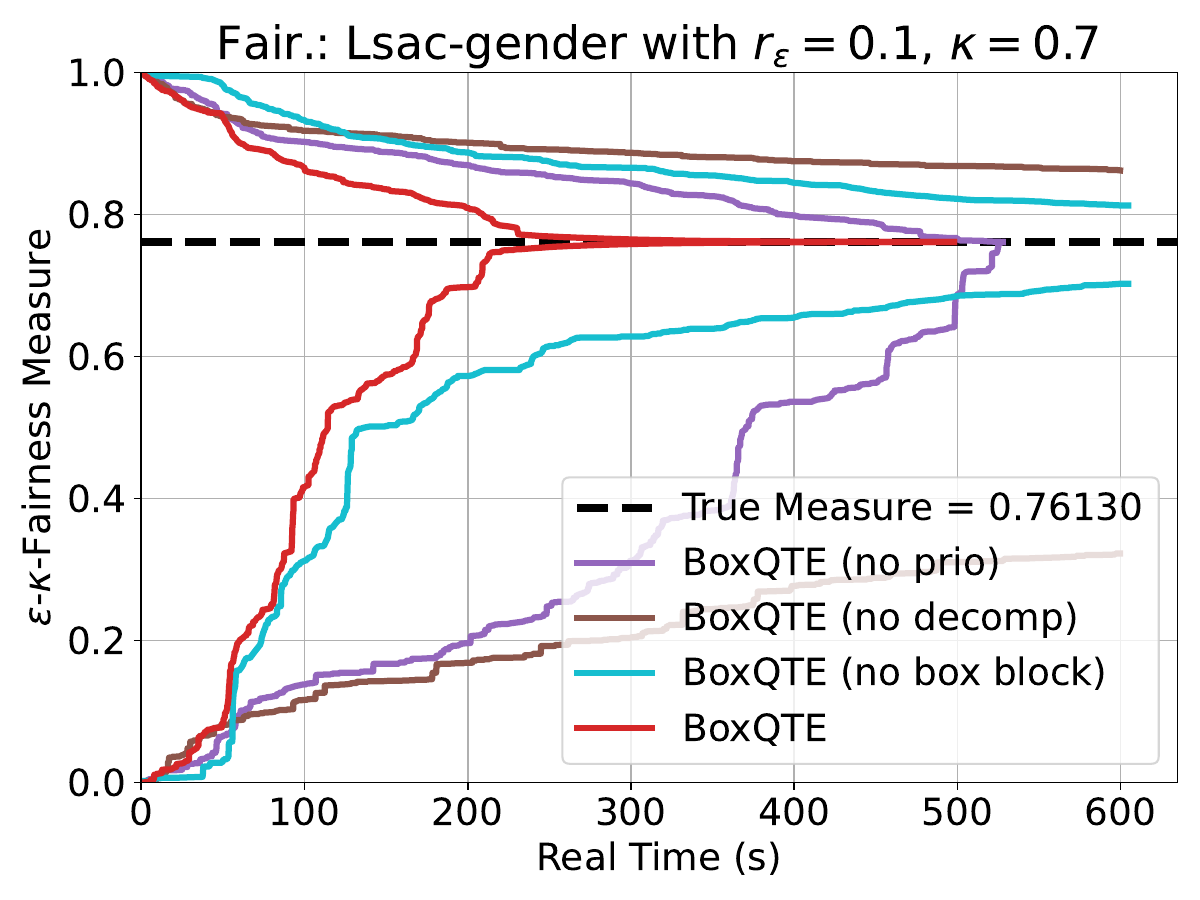}
        %\caption{1}
        %\label{fig:image1}
    \end{subfigure}
    \hfill
    \begin{subfigure}[b]{0.24\textwidth}
        \centering
        \includegraphics[width=\textwidth]{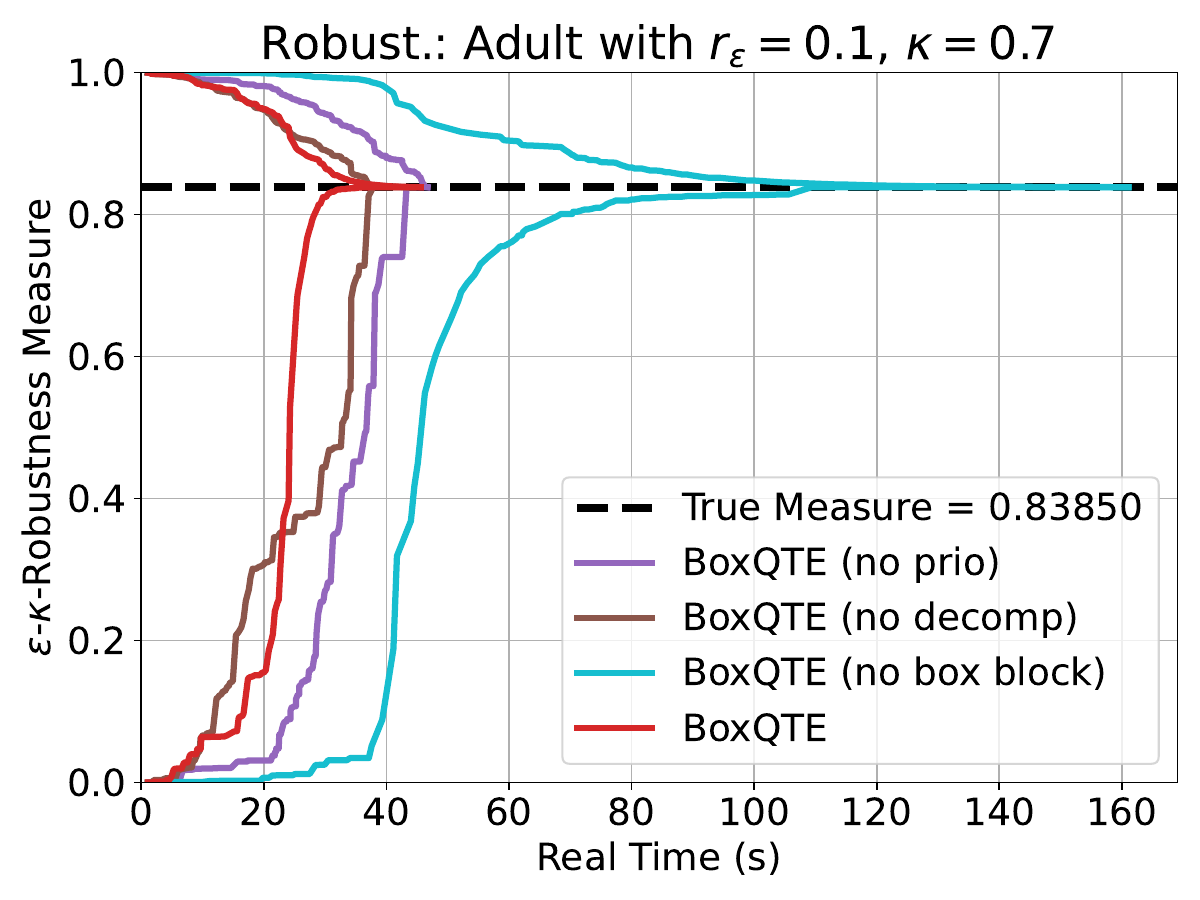}
        %\caption{1}
        %\label{fig:image2}
    \end{subfigure}
    \hfill
    \begin{subfigure}[b]{0.24\textwidth}
        \centering
        \includegraphics[width=\textwidth]{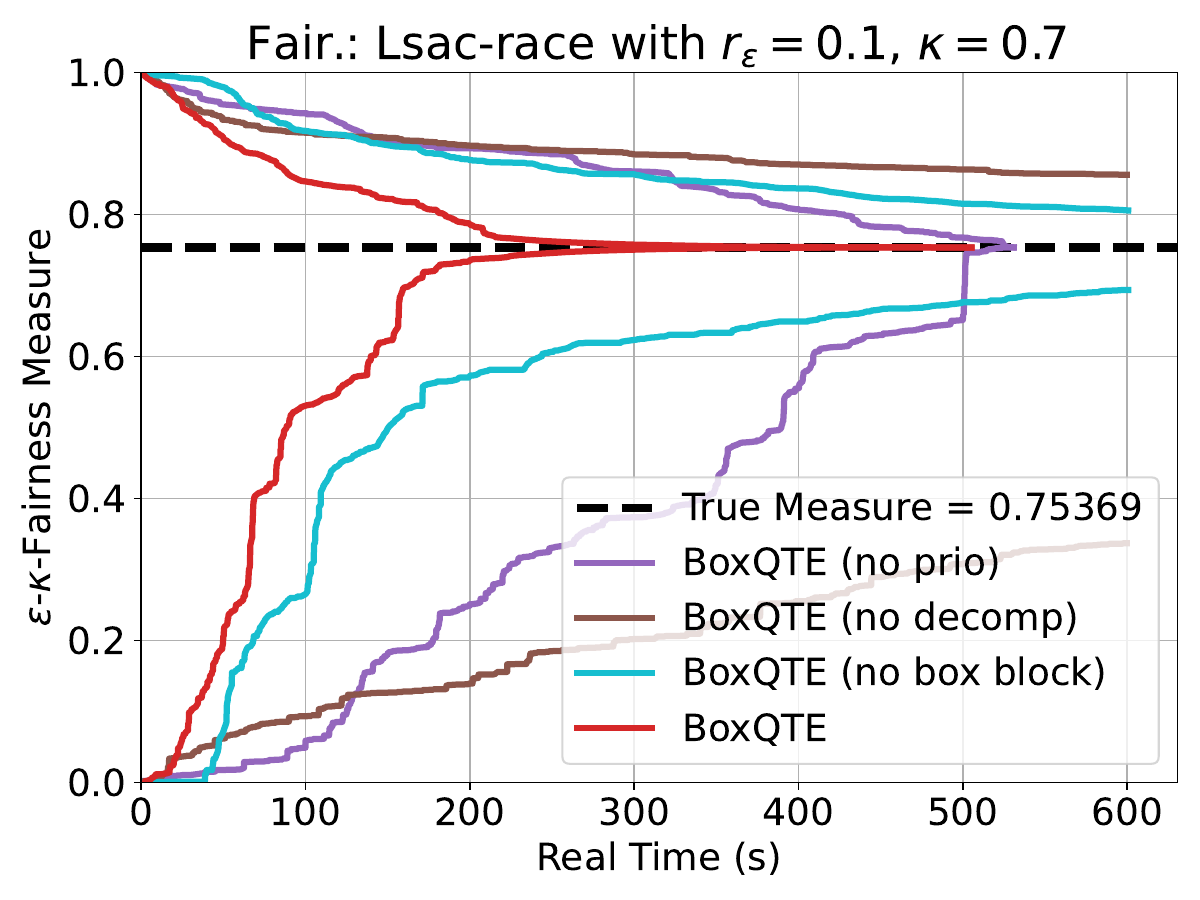}
        %\caption{1}
        %\label{fig:image2}
    \end{subfigure}
    \hfill
    \begin{subfigure}[b]{0.24\textwidth}
        \centering
        \includegraphics[width=\textwidth]{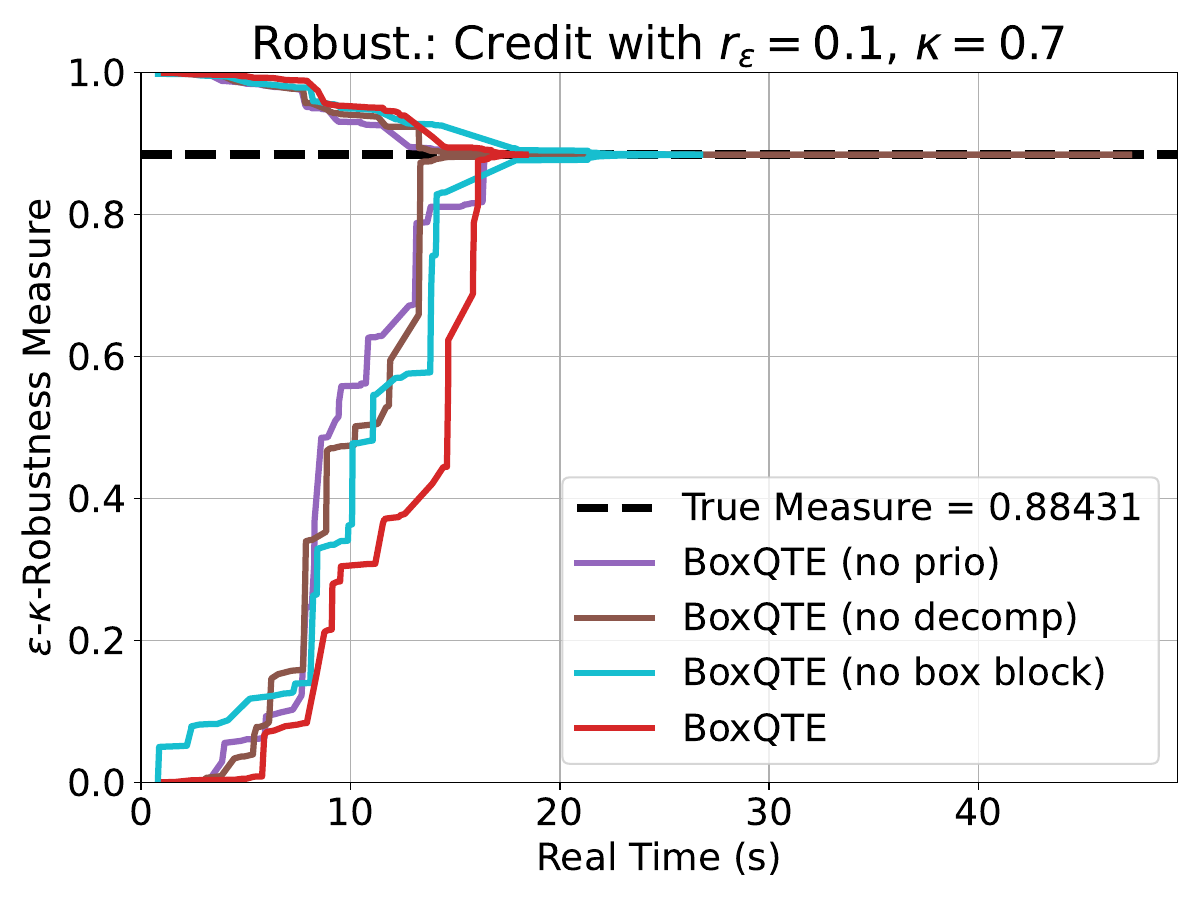}
        %\caption{1}
        %\label{fig:image2}
    \end{subfigure}
  \caption{Evolution of bounds over time of \boxmethod{} and its three enhancements with time limit of 600s.}
  \label{fig:res_evolution_box_vs_imporvement}
\end{figure*}

\subsubsection{RQ2: \RQtwo{}}
We compare \boxmethod{} with its three modified variants, \boxmethod{ (no-prior)}, \boxmethod{ (no-decomp)}, and \boxmethod{ (no-box-block)}, under a setting with $\epsr{}=0.1$, $\kappa=0.7$.
\prettyref{tab:res_box_vs_improvement} summarizes their quantification results on $15$ configurations, and \prettyref{fig:res_evolution_box_vs_imporvement} shows the evolution of bounds over time.
\boxmethod{} converges on all $15$ configurations with an average runtime of 129.4s.
\boxmethod{ (no-box-block)} converges on $12$ out of $15$ configurations. For the remaining three, their final bound widths (`Gap') is $0.11$. Its average runtime is 184.4s, which is $1.42$ times than \boxmethod{}, indicating that the box blocking enhancement contributes to efficiency.
\boxmethod{ (no-decomp)} also converges on $12$ configurations. For the three incomplete cases, the final `Gaps' are $0.52$, $0.54$, and $0.51$. Its average runtime is 211.0s, which is $1.63$ times than \boxmethod{}. This suggests that task decomposition, in conjunction with box blocking, plays a more substantial role in efficiency.
\boxmethod{ (no-prior)} converges on all $15$ configurations, with an average runtime of 135.3s, close to the \boxmethod{}. This indicates that the task priority enhancement has minimal impact on overall efficiency. However, as shown in \prettyref{fig:res_evolution_box_vs_imporvement}, task priority contributes to faster bound tightening in the early stages. For example, in the leftmost plot of \prettyref{fig:res_evolution_box_vs_imporvement}, the `Gap' at 250s is $0.014$ for \boxmethod{}, compared to $0.605$ for \boxmethod{ (no-prior)}. We also observe that this effect becomes more pronounced in longer-running quantifications.
For users who do not require exact results, this early-stage narrowing of the bounds enabled by task priority may offer practical advantages.

\begin{tcolorbox}
[size=small, arc=0pt, boxsep=5pt,left=2pt,right=2pt,top=0pt,bottom=0pt]
\textbf{Answer for RQ2:}
    The box blocking and task decomposition enhancements contribute to the overall efficiency of \boxmethod{}, while the task priority enhancement facilitates faster bound tightening in the early stages and becomes more effective in longer quantifications.
\end{tcolorbox}

\begin{table}[t]
\caption{Runtime and \#IDIs generated by \boxmethod{} and state-of-the-art fairness testing techniques with $\epsr{}=0.0$, $\kappa=0.5$, and time limit of 3600s.}
\label{tab:res_box_vs_sota_fairness_testing}
\centering
\setlength{\tabcolsep}{4pt}
\renewcommand{\arraystretch}{1.2}
\resizebox{1\columnwidth}{!}{
\begin{tabular}{c|c||c|c|rrr|rrrr|r}
\hline 
\multirow{3}{*}{Dataset } & \multirow{3}{*}{Attr. } & \multicolumn{5}{c|}{\boxmethod{} } & \multicolumn{5}{c}{Fairness Testing Methods}\tabularnewline
\cline{3-12} \cline{4-12} \cline{5-12} \cline{6-12} \cline{7-12} \cline{8-12} \cline{9-12} \cline{10-12} \cline{11-12} \cline{12-12} 
 &  & \multirow{2}{*}{Measure} & \multirow{2}{*}{\#IDIs} & \multicolumn{3}{c|}{Time} & \multicolumn{4}{c|}{\#IDIs} & \multirow{2}{*}{Time}\tabularnewline
 &  &  &  & Quant.  & Samp.  & Total  & \aft{}  & \expga{}  & \themis{}  & \sg{}  & \tabularnewline
\hline 
\hline 
\multirow{3}{*}{Adult } & age  & 0.9964  & \multicolumn{1}{r|}{\textbf{1000000} } & 26.4  & 3.5  & \textbf{29.9} & 94328  & 6801  & 91  & 309  & 3600\tabularnewline
 & race  & 1.0000  & \multicolumn{1}{r|}{0 } & <0.1  & 0.0  & <0.1 & 0  & 0  & 0  & 0  & 3600\tabularnewline
 & gender  & 1.0000  & \multicolumn{1}{r|}{0 } & <0.1  & 0.0  & <0.1 & 0  & 0  & 0  & 0  & 3600\tabularnewline
\hline 
Bank  & age  & 0.9183  & \multicolumn{1}{r|}{\textbf{1000000 }} & 101.9  & 4.3  & \textbf{106.2} & 155892  & 6942  & 2257  & 1744  & 3600\tabularnewline
\hline 
\multirow{2}{*}{COMPAS} & age  & 0.9715  & \multicolumn{1}{r|}{\textbf{31658 }} & 34.0  & 1.8  & \textbf{35.8} & 15754  & 105  & 1140  & 98  & 3600\tabularnewline
 & gender  & 0.9854  & \multicolumn{1}{r|}{\textbf{8118 }} & 33.5  & 1.5  & \textbf{35.0} & 3793  & 319  & 1678  & 113  & 3600\tabularnewline
\hline 
\multirow{2}{*}{Credit } & age  & 0.9996  & \multicolumn{1}{r|}{\textbf{999987 }} & 27.0  & 4.7  & \textbf{31.7} & 66  & 1930  & 12  & 29  & 3600\tabularnewline
 & gender  & 0.9781  & \multicolumn{1}{r|}{\textbf{999997 }} & 31.2  & 4.9  & \textbf{36.1} & 142362  & 45311  & 2455  & 649  & 3600\tabularnewline
\hline 
\multirow{2}{*}{LSAC } & age  & 0.9991  & \multicolumn{1}{r|}{\textbf{999999 }} & 226.3  & 3.3  & \textbf{229.5} & 0  & 18481  & 75  & 0  & 3600\tabularnewline
 & gender  & 1.0000  & \multicolumn{1}{r|}{0 } & <0.1  & 0.0  & <0.1 & 0  & 0  & 0  & 0  & 3600\tabularnewline
\hline 
\hline 
\multicolumn{2}{c||}{Avg. } & ---  & \multicolumn{1}{r|}{\textbf{503976} } & 48.0  & 2.4  & \textbf{50.4 } & 41219  & 7989  & 771  & 294  & 3600\tabularnewline
\hline 
\end{tabular}
}
\end{table}

\subsubsection{RQ3: \RQthree{}}
Efficiently identifying a set of \emph{individual discriminatory instances} (or \emph{IDIs}), i.e., counterexamples to fairness, is a key objective in fairness testing\cite{Chen+TOSEM24,Fan+ICSE22,zhao+ASE24}.
Our quantification method, \boxmethod{}, can captures the entire counterexample space, thus it can be used in fairness testing by sampling input from counterexample space.
Since existing fairness tesing methods cannot support $\kappa$ and $\vec{\epsilon}$, all experiments are under the settings of $\epsr{}=0$ and $\kappa=0.5$.
After quantification, \boxmethod{} randomly samples $1,000,000$ inputs as IDIs.
\prettyref{tab:res_box_vs_sota_fairness_testing} reports the resutls of runtime and the number of identified IDIs (\#IDIs) across ten configurations.
On average, \boxmethod{} takes only $1.4\%$ of the time used by other fairness testing methods (50.4s vs. 3600s) and detects significantly more IDIs: $12.2$ times to \aft{}, $63.0$ times to \expga{}, $653.6$ times to \themis, and $1714.2$ times to \sg{}.
Notably, for challenging configurations like `Credit-age' with a high fairness measure ($0.9996$), while the best competing method finds at most $1930$ IDIs within 3600s, BoxQTE identifies $999,987$ within 31.7s.

\begin{tcolorbox}
[size=small, arc=0pt, boxsep=5pt,left=2pt,right=2pt,top=0pt,bottom=0pt]
\textbf{Answer for RQ3:}
    When used for fairness testing, \boxmethod{} can detect more individual discriminatory instances in substantially less time than state-of-the-art fairness testing techniques.
\end{tcolorbox}

\section{Discussion}
\label{sec:discussion}

\paragraph{Extensions to multi-class tree ensembles}
Our method can be directly extended to multi-class RFs since its confidence computation does not involve any nonlinear functions. For multi-class GBDTs, however, additional handling is required due to their nonlinear softmax functions. One straightforward solution is to adopt the approximation technique used in \cite{Athavale+CAV24}, which approximates the softmax function using piecewise linear functions and the max function, both of which are compatible with SMT encoding. 

\paragraph{Scalability}
Although our proposed method has proved effective and superior to existing baselines on the five standard fairness-oriented datasets, enhancing its scalability remains an important direction for future work. The computational burden of our method increases steeply with the size of a tree ensemble because it is proportional to the number of box pairs to be examined, and the number of boxes typically grows exponentially.

\paragraph{Integration with approximation-guided fairness testing.}
Several black-box fairness testing methods such as \aft{}~\cite{zhao+ASE24}, \sg{}~\cite{Aggarwal+FSE19}, and \vbt{}~\cite{Sharma+ISSTA20}  use an approximation-guided strategy: they train a decision tree surrogate for the target model and sample counterexamples from this surrogate to uncover fairness violations. Although \aft{} outperforms many peers, a single tree often fails to capture the complex models such as DNNs~\cite{zhao+ASE24}. Adopting a more complex tree ensemble surrogate and pairing it with our quantification method may generate test cases more efficiently and improve approximation-based fairness testing.

\section{Conclusion}
\label{sec:conclusion-and-future-work}
In this work, we introduce \boxmethod{}, an quantative verification framework for fairness in tree ensembles.
By exploiting the discrete, box-structured partitioning of the input space in tree ensembles, \boxmethod{} can enumerate path-tuple regions and obtain upper and lower bounds for quantitative fairness measures.
Experimental results across multiple benchmarks demonstrate that \boxmethod{} significantly outperforms the existing \cegarmethod{} approach.
Future work will focus on improving scalability for larger tree ensembles and integrating these quantification results into model debugging and bias-mitigation pipelines.

\begin{acks}
This paper is supported by JST SPRING, Grant Number JPMJSP2131.
\end{acks}

\bibliographystyle{ACM-Reference-Format}
\bibliography{references}

\clearpage
\appendix

\renewcommand{\thetheorem}{\Alph{section}.\arabic{theorem}}
\renewcommand{\thelemma}{\Alph{section}.\arabic{lemma}}
\renewcommand{\thedefinition}{\Alph{section}.\arabic{definition}}
\renewcommand{\thefigure}{\Alph{section}.\arabic{figure}}
\renewcommand{\thetable}{\Alph{section}.\arabic{table}}
\renewcommand{\thealgocf}{\Alph{section}.\arabic{algocf}}

\numberwithin{algocf}{section}

% reset counter
\setcounter{equation}{0}

\setcounter{theorem}{0}
\setcounter{figure}{0}
\setcounter{table}{0}
\setcounter{definition}{0}
\setcounter{lemma}{0}

\section{Definitions of Robustness and Fairness\label{app:definition-robustness-fairness}}
We review the definitions of several basic concepts of robustness and fairness. 

\emph{Local robustness} requires that for a given input, small perturbations to it do not change its classification outcome.
\begin{definition}[Local Robustness]
% Local rubstness 
Given a tuple of real numbers  $\vec{\epsilon} = (\epsilon_1, \dots, \epsilon_n)$ of size $n$, where $\epsilon_i>0$, and an input $\vec{x}$, a classifier $f$ is \emph{locally $\epsilon$-robust} at $\vec{x}$ if the following condition is satisfied:
\begin{equation*}
\begin{aligned}
\forall \vec{x'} \in X, \bigwedge_{i=1}^{n} |x_i - x_i'| \leq \epsilon_i \to f(\vec{x}) = f(\vec{x'}).
\end{aligned}
\end{equation*}
\end{definition}
However, as it is input-specific, local robustness cannot guarantee robustness for other inputs.
\emph{Global robustness}, a natural generalization of local robustness, requires that perturbations do not affect classification outcomes for any input across the entire input space.
\begin{definition}[Global Robustness]
Given a tuple of real numbers $\vec{\epsilon} = (\epsilon_1, \dots, \epsilon_n)$ of size $n$, where $\epsilon_i>0$, a classifier $f$ is \emph{globally $\epsilon$-robust} if the following condition is satisfied:
\begin{equation*}
\begin{aligned}
\forall \vec{x}, \vec{x'} \in X, \bigwedge_{i=1}^{n} |x_i - x_i'| \leq \epsilon_i \to f(\vec{x}) = f(\vec{x'}).
\end{aligned}
\end{equation*}
\end{definition}

The field of ML fairness is still evolving, with various formulations proposed to address different aspects of fairness.
\cite{fairify+biswas+ICSE23} recently introduced a fairness formulation named \emph{$\epsilon$-fairness} that requires the following: for any input, altering its \emph{sensitive attributes} (such as gender, race, and age) while applying perturbations to its non-sensitive attributes should not result in a change in the classification outcome.
\emph{Individual fairness} is a special case of $\epsilon$-Fairness by setting $\epsilon$ to $0$.
\begin{definition}[Individual Fairness]
Given an index set of sensitive attributes $S \subseteq A$, a classifier $f$ is \emph{individually fair} with respect to $S$ if the following condition is satisfied:
\begin{equation*}
\begin{aligned}
\forall \vec{x}, \vec{x'} \in X, \bigwedge_{i \in A \setminus S} x_i = x'_i \land \bigvee_{j \in  S} x_j \neq x'_j \to f(\vec{x}) = f(\vec{x'}).
\end{aligned}
\end{equation*}
\end{definition}

\begin{definition}[$\epsilon$-Fairness]
Given a tuple of real numbers $\vec{\epsilon} = (\epsilon_1, \dots, \epsilon_n)$, where $\epsilon_i>0$, and an index set of sensitive attributes $S \subseteq A$, a classifier $f$ is \emph{$\epsilon$-fair} with respect to $S$ if the following condition is satisfied:
\begin{equation*}
\begin{aligned}
\forall \vec{x}, \vec{x'} \in X, \bigwedge_{i \in A \setminus 
 S} |x_i - x'_i| \leq \epsilon_i \land \bigvee_{j \in S} x_j \neq x'_j \to f(\vec{x}) = f(\vec{x'}).
\end{aligned}
\end{equation*}
\end{definition}

\section{An example of SMT encoding}
\prettyref{fig:example-smt} shows an example of SMT formula for verifying fairness of the GBDT tree ensemble classifier in \prettyref{fig:tree-ensemble}.

\begin{figure}[t]
    \centering
    \begin{lstlisting}[mathescape=true]
    // Declaration of variables with their domains
    $0 \leq x_1 \leq 1000$;  $0 \leq x_1' \leq 1000$
    $0 \leq x_2 \leq 4$;  $0 \leq x_2' \leq 4$
    $0 \leq x_3 \leq 100$;  $0 \leq x_3' \leq 100$
 
    // Constraints obtained by instantiating $\Phi{\text{trees}}(\vec{x}, \vec{v}, \vec{p})$ with $\vec{x}, \vec{v}, \vec{p}$ :
    $(0 \leq x_1 < 300) \And (0 \leq x_2 < 2) \And (0 \leq x_3 \leq 100)  \rightarrow (v_1 = -0.17) \And (p_1 = 0)$
    $(300 \leq x_1 < 1000) \And (0 \leq x_2 < 2) \And (0 \leq x_3 \leq 100)  \rightarrow (v_1 = 0.85) \And (p_1 = 1)$
    $(0 \leq x_1 \leq 1000) \And (2 \leq x_2 \leq 4) \And (0 \leq x_3 \leq 100)  \rightarrow (v_1 = -0.02) \And (p_1 = 2)$
    $(0 \leq x_1 < 600) \And (0 \leq x_2 < 3) \And (0 \leq x_3 \leq 100)  \rightarrow (v_2 = 0.7) \And (p_2 = 3)$
    $(0 \leq x_1 < 600) \And (3 \leq x_2 \leq 4) \And (0 \leq x_3 \leq 100)  \rightarrow (v_2 = 1.62) \And (p_2 = 4)$
    $(600 \leq x_1 \leq 1000) \And (0 \leq x_2 \leq 4) \And (0 \leq x_3 < 50)  \rightarrow (v_2 = -0.19) \And (p_2 = 5)$
    $(600 \leq x_1 \leq 1000) \And (0 \leq x_2 \leq 4) \And (50 \leq x_3 \leq 100)  \rightarrow (v_2 = 1.75) \And (p_2 = 6)$
    $(0 \leq x_1 \leq 1000) \And (0 \leq x_2 \leq 4) \And (0 \leq x_3 < 50)  \rightarrow (v_3 = 0.72) \And (p_3 = 7)$
    $(0 \leq x_1 \leq 1000) \And (0 \leq x_2 \leq 4) \And (50 \leq x_3 \leq 100)  \rightarrow (v_3 = -1.3) \And (p_3 = 8)$
    $\text{leaf\_sum} = v_1 + v_2 + v_3$
    // Constraints obtained by instantiating $\Phi{\text{trees}}(\vec{x}, \vec{v}, \vec{p})$ with $\vec{x'}, \vec{v'}, \vec{p'}$ :
    $(0 \leq x_1' < 300) \And (0 \leq x_2' < 2) \And (0 \leq x_3' \leq 100)  \rightarrow (v_1' = -0.17) \And (p_1' = 0)$ 
    ...
    $\text{leaf\_sum}' = v'_1 + v'_2 + v'_3$

    // $\Phi{\kappa}(\vec{v}, \kappa)$:
    $\text{leaf\_sum} > 1.3862943611198908$  // $\text{logit}(0.8) = 1.3862943611198908$
    
    // $\Phi{\text{unfair}}(\vec{x}, \vec{x'}, \vec{v}, \vec{v'}, \vec{\epsilon})$:
    $(-5 \leq x_1 - x'_1 \leq 5) \And (x_2 \neq x'_2) \And (-5 \leq x_3 - x'_3 \leq 5)$ // $\epsilon_1 = \epsilon_3 = 5$
    $(leaf\_sum<0 \And leaf\_sum' > 0) \Or (leaf\_sum > 0 \And leaf\_sum'< 0)$
    \end{lstlisting}
\caption{SMT formula $\Phi_{\text{unfair}}$ in a readable form to verify $\epsilon$-$\kappa$-fairness for the GBDT tree ensemble classifier  in \prettyref{fig:tree-ensemble}.
$x_1$, $x_2$ and $x_3$ correspond to the attribute of `income', `race', and `age', respectively.
The attribute `race' is designated as the sensitive attribute.
The tolerance and confidence are set as follows: $\epsilon_1 = \epsilon_3 = 5$ and $\kappa = 0.8$.
In practical implementation, each path can be uniquely represented by a natural number.
\label{fig:example-smt}}
\end{figure}

\end{document}